\documentclass[final]{cvpr}

\usepackage{times}
\usepackage{epsfig}
\usepackage{graphicx}
\usepackage{amsmath}
\usepackage{amssymb}
\usepackage{subcaption}
\usepackage{multirow}
\usepackage{algorithm, algpseudocode}

\usepackage[pagebackref=true,breaklinks=true,colorlinks,bookmarks=false]{hyperref}

\def\cvprPaperID{6930} 
\def\confYear{CVPR 2021}
\DeclareMathOperator*{\argmax}{\arg\!\max}
\newtheorem{definition}{Definition}

\begin{document}

\title{Enhancing the Transferability of Adversarial Attacks through Variance Tuning}

\author{Xiaosen Wang \quad Kun He \thanks{Corresponding author.}\\
School of Computer Science and Technology, Huazhong University of Science and Technology\\
{\tt\small \{xiaosen,brooklet60\}@hust.edu.cn}
}

\maketitle

\pagestyle{empty}
\thispagestyle{empty}

\begin{abstract}
Deep neural networks are vulnerable to adversarial examples that mislead the models with imperceptible perturbations. Though adversarial attacks have achieved incredible success rates in the white-box setting, most existing adversaries often exhibit weak transferability in the black-box setting, especially under the scenario of attacking models with defense mechanisms. In this work, we propose a new method called variance tuning to enhance the class of iterative gradient based attack methods and improve their attack transferability. Specifically, at each iteration for the gradient calculation, instead of directly using the current gradient for the momentum accumulation, we further consider the gradient variance of the previous iteration to tune the current gradient so as to stabilize the update direction and escape from poor local optima. Empirical results on the standard ImageNet dataset demonstrate that our method could significantly improve the transferability of gradient-based adversarial attacks. Besides, our method could be used to attack ensemble models or be integrated with various input transformations. Incorporating variance tuning with input transformations on iterative gradient-based attacks in the multi-model setting, the integrated method could achieve an average success rate of 90.1\% against nine advanced defense methods, improving the current best attack performance significantly by 85.1\%. Code is available at \url{https://github.com/JHL-HUST/VT}. 
\end{abstract}

\section{Introduction}
Deep Neural Networks (DNNs) are known to be vulnerable to adversarial examples \cite{szegedy2014intriguing, goodfellow2015FGSM}, which are indistinguishable from legitimate ones by adding small perturbations, but lead to incorrect model prediction. In recent years, it has garnered an increasing interest to craft adversarial examples \cite{moosavi2016deepfool, carlini2017cw, kurakin2017IFGSM, athalye2018obfuscated, li2019nattack}, because it not only can identify the model vulnerability \cite{carlini2017cw, athalye2018obfuscated}, but also can help improve the model robustness \cite{goodfellow2015FGSM, madry2018pgd, tramer2018ensemble, zhang2019theoretically}. 
Moreover, adversarial examples also exhibit good transferability across the models \cite{papernot2017practical, liu2017delving}, \ie the adversaries crafted for one model can still fool other models, which enables black-box attacks in the real-world applications without any knowledge of the target model. 

\begin{figure}
    \centering
    \begin{minipage}[c]{0.14\textwidth} 
        \begin{subfigure}{\textwidth}
          \centering 
          \includegraphics[width=\linewidth]{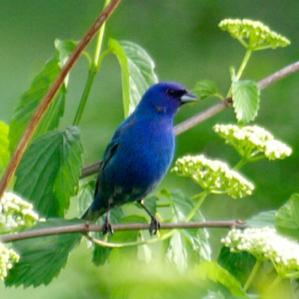}
          \vspace{-1.7em}
          \caption*{Raw Image}
        \end{subfigure}
    \end{minipage}
    \hspace{0.1cm}
    \begin{minipage}[c]{0.14\textwidth} 
        \begin{subfigure}{\textwidth}
          \centering 
          \includegraphics[width=\linewidth]{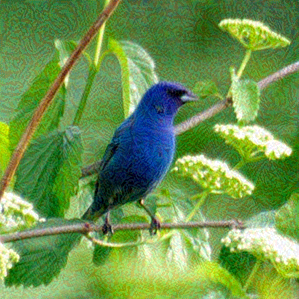}
          \vspace{-1.7em}
          \caption*{MI-FGSM}
        \end{subfigure}\\
        \begin{subfigure}{\textwidth} 
          \centering 
          \includegraphics[width=\linewidth]{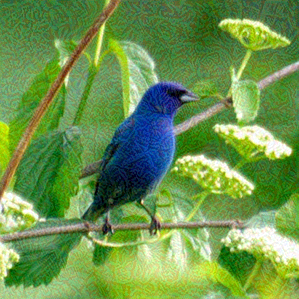}
          \vspace{-1.7em}
          \caption*{NI-FGSM}
        \end{subfigure}%
    \end{minipage}
     \hspace{0.1cm}
    \begin{minipage}[c]{0.14\textwidth} 
        \begin{subfigure}{\textwidth}
          \centering 
          \includegraphics[width=\linewidth]{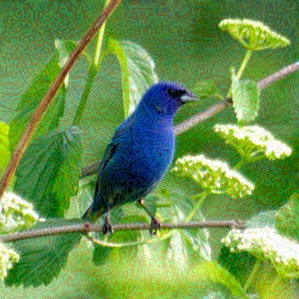}
          \vspace{-1.7em}
          \caption*{VMI-FGSM}
        \end{subfigure}\\
        \begin{subfigure}{\textwidth} 
          \centering 
          \includegraphics[width=\linewidth]{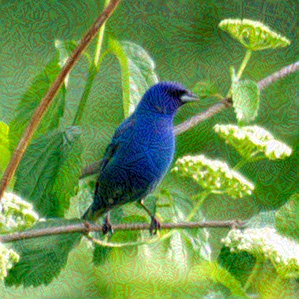}
          \vspace{-1.7em}
          \caption*{VNI-FGSM}
        \end{subfigure}%
    \end{minipage}
    \vspace{-0.4em}
    \caption{Adversarial examples crafted by MI-FGSM~\cite{dong2018boosting}, NI-FGSM \cite{lin2020nesterov}, the proposed VMI-FGSM and VNI-FGSM on the Inc-v3 model \cite{szegedy2016inceptionv3} with the maximum perturbation of $\epsilon = 16$. VMI-FGSM and VNI-FGSM generate visually similar adversaries as other attacks but lead to much higher transferability.}
    \label{fig:adv_images}
    \vspace{-0.6em}
\end{figure}

In the white-box setting that the attacker can access the architecture and parameters of the target model, existing adversarial attacks have exhibited great effectiveness \cite{kurakin2017IFGSM, carlini2017cw, athalye2018obfuscated} but with low transferability, especially for models equipped with defense mechanisms \cite{tramer2018ensemble, xie2018mitigating,liao2018defense, naseer2020NRP}. To address this issue, recent works focus on improving the transferability of adversarial examples by advanced gradient calculation (\eg Momentum, Nesterov's accelerated gradient, 
\etc) \cite{dong2018boosting, lin2020nesterov},
attacking multiple models \cite{liu2017delving}, or adopting various input transformations (\eg random resizing and padding, translation, scale, admix, \etc) \cite{xie2019improving, dong2019evading, lin2020nesterov, wang2021Admix}. However, there still exists a big gap between white-box attacks and transfer-based black-box attacks with regard to attack performance.

In this work, we propose 
a novel variance tuning iterative gradient-based method to enhance the transferability of the generated adversarial examples. Different from existing gradient-based methods that perturb the input in the gradient direction of the loss function, or momentum iterative gradient-based methods that accumulate a velocity vector in the gradient direction, at each iteration our method additionally tunes the current gradient with the gradient variance in the neighborhood of the previous data point. The key idea is to reduce the variance of the gradient at each iteration so as to stabilize the update direction and escape from poor local optima during the search process.
Empirical results on the standard ImageNet dataset demonstrate that, compared with state-of-the-art momentum-based adversarial attacks \cite{dong2018boosting, lin2020nesterov}, the proposed method could achieve significantly higher success rates for black-box models, meanwhile maintain similar success rates for white-box models. For instance, the proposed method improves the success rates of the momentum based attack \cite{dong2018boosting} for more than $20\%$ using adversarial examples generated on the Inc-v3 model \cite{szegedy2016inceptionv3}. The adversarial examples crafted by various attacks are illustrated in Figure~\ref{fig:adv_images}.

To further demonstrate the effectiveness of our method, we combine  variance tuning with several gradient-based attacks for ensemble models \cite{liu2017delving} and integrate these attacks with various input transformations \cite{xie2019improving, dong2019evading, lin2020nesterov}. Extensive experiments show that our integrated method could remarkably improve the attack transferability. In addition, we compare our attack method with the state-of-the-art attack methods \cite{dong2018boosting,xie2019improving,dong2019evading,lin2020nesterov} against nine advanced defense methods \cite{liao2018defense, xie2018mitigating, xu2018BitReduction, guo2018countering, liu2019FD, jia2019comdefend, cohen2019certified, naseer2020NRP}. Our integrated method yields an average success rate of $67.0\%$, which outperforms the baselines by a large margin of $17.5\%$ in the single model setting, and an average success rate of $90.1\%$, which outperforms the baselines by a clear margin of $6.6\%$ in the multi-model setting.

\section{Related Work}
Let $x$ be a benign image, $y$ the corresponding true label and $f(x;\theta)$ the classifier with parameters $\theta$ that outputs the prediction result. Let $J(x,y;\theta)$ denote the loss function of classifier $f$ (\eg the cross-entropy loss). 
We define the adversarial attack as finding an example $x^{adv}$ that satisfies $\|x-x^{adv}\|_p < \epsilon$ but misleads the model prediction, $f(x;\theta)\neq f(x^{adv};\theta)$. Here $\|\cdot\|_p$ denotes the $p$-norm distance and we focus on $p=\infty$ to align with previous works.

\subsection{Adversarial Attacks}
Numerous adversarial attack methods have been proposed in recent years, including gradient-based methods \cite{goodfellow2015FGSM, kurakin2017IFGSM, madry2018pgd, dong2018boosting,lin2020nesterov}, optimization-based methods \cite{szegedy2014intriguing, carlini2017cw}, score-based methods \cite{ilyas2018black, li2019nattack} and decision-based methods \cite{brendel2018decision, chen2020boosting}. In this work, we mainly focus on the attack transferability and provide a brief overview on two branches of transfer-based attacks in this subsection.

\subsubsection{Gradient-based Attacks} 
The first branch focuses on improving the transferability of gradient-based attacks by advanced gradient calculation. 

\textbf{Fast Gradient Sign Method (FGSM).} FGSM \cite{goodfellow2015FGSM} is the first gradient-based attack that crafts an adversarial example $x^{adv}$ by maximizing the loss function $J(x^{adv}, y; \theta)$ with a one-step update:
\begin{equation*}
    x^{adv}=x+\epsilon \cdot \text{sign}(\nabla_x J(x, y; \theta)),
\end{equation*}
where $\nabla_x J(x, y; \theta)$ is the gradient of loss function \wrt $x$ and $\text{sign}(\cdot)$ denotes the sign function.

\textbf{Iterative Fast Gradient Sign Method (I-FGSM).} I-FGSM \cite{kurakin2017IFGSM} extends FGSM to an iterative version with a small step size $\alpha$:
\begin{gather}
\label{eq:ifgsm}
    x_{t+1}^{adv} = x_t^{adv} + \alpha \cdot \text{sign}(\nabla_{x_t^{adv}} J(x_t^{adv}, y; \theta)), \\
    x_0^{adv} = x. \nonumber
\end{gather}

\textbf{Momentum Iterative Fast Gradient Sign Method (MI-FGSM).} MI-FGSM \cite{dong2018boosting} integrates the momentum into I-FGSM and achieves much higher transferability: 
\begin{gather}
    g_{t+1} = \mu \cdot g_t + \frac{\nabla_{x_t^{adv}}J(x_t^{adv}, y; \theta)}{\|\nabla_{x_t^{adv}}J(x_t^{adv}, y; \theta)\|_1}, \label{eq:momentum}\\
    x_{t+1}^{adv} = x_t^{adv} + \alpha \cdot \text{sign}(g_{t+1}), \nonumber
\end{gather}
where $g_0 = 0$ and $\mu$ is the decay factor.

\textbf{Nesterov Iterative Fast Gradient Sign Method (NI-FGSM).} NI-FGSM~\cite{lin2020nesterov} adopts Nesterov's accelerated gradient \cite{Nesterov1983}, and substitutes $x_t^{adv}$ in Eq. \eqref{eq:momentum} with $x_t^{adv} + \alpha \cdot \mu \cdot g_t$ to further improve the transferability of MI-FGSM.

\subsubsection{Input Transformations} 
The second branch focuses on adopting various input transformations to enhance the attack transferability.

\textbf{Diverse Input Method (DIM).} DIM \cite{xie2019improving} applies random resizing and padding to the inputs with a fixed probability, and feeds the transformed images into the classifier for the gradient calculation to improve the transferability.

\textbf{Translation-Invariant Method (TIM).} TIM \cite{dong2019evading} adopts a set of images to calculate the gradient, which performs well especially for black-box models with defense mechanisms. To reduce the calculation on gradients, Dong \etal \cite{dong2019evading} shift the image within small magnitude and approximately calculate the gradients by convolving the gradient of untranslated images with a kernel matrix.

\textbf{Scale-Invariant Method (SIM).} SIM \cite{lin2020nesterov} introduces the scale-invariant property and calculates the gradient over a set of images scaled by factor $1/2^i$ on the input image to enhance the transferability of the generated adversarial examples, where $i$ is a hyper-parameter.

Note that different input transformations, DIM, TIM and SIM, can be naturally integrated with  gradient-based attack methods. Lin \etal \cite{lin2020nesterov} have shown that the combination of these methods, denoted as Composite Transformation Method (CTM), is the current strongest transfer-based black-box attack method. 
In this work, the proposed variance tuning based method aims to improve the transferability of the gradient-based attacks (\eg MI-FGSM, NI-FGSM), and can be combined with various input transformations to further improve the attack transferability.

\subsection{Adversarial Defenses}
In contrast, to mitigate the threat of adversarial examples, various adversarial defense methods have been proposed. One promising defense method is called \textit{adversarial training} \cite{goodfellow2015FGSM, kurakin2017adversarial, madry2018pgd} that injects the adversarial examples into the training data to improve the model robustness. Tram{\`e}r \etal \cite{tramer2018ensemble} propose \textit{ensemble adversarial training} by augmenting the training data with perturbations transferred from several models, which can further improve the robustness against transfer-based black-box attacks. However, such adversarial training methods, as one of the most powerful and extensively investigated defense methods, often result in high computation cost and are difficult to be scaled to large-scale datasets and complex neural networks~\cite{kurakin2017adversarial}.

Guo \etal \cite{guo2018countering} utilize a set of image transformations (\eg JPEG compression, Total Variance Minimization, \etc) on the inputs to eliminate adversarial perturbations before feeding the images to the models. Xie \etal \cite{xie2018mitigating} adopt random resizing and padding (R\&P) on the inputs to mitigate the adversarial effect. Liao \etal \cite{liao2018defense} propose to train a high-level representation denoiser (HGD) to purify the input images. Xu \etal \cite{xu2018BitReduction} propose two feature squeezing methods: bit reduction (Bit-Red) and spatial smoothing to detect adversarial examples. Feature distillation (FD) \cite{liu2019FD} is a JPEG-based defensive compression framework against adversarial examples. ComDefend \cite{jia2019comdefend} is an end-to-end image compression model to defend adversarial examples. Cohen \etal \cite{cohen2019certified} adopt randomized smoothing (RS) to train a certifiably robust ImageNet classifier. Naseer \etal \cite{naseer2020NRP} design a neural representation purifier (NRP) model that learns to purify the adversarially perturbed images based on the automatically derived supervision.

\section{Methodology}
For the methodology section, we first introduce our motivation, then provide a detailed description of the proposed method. In the end, we formulize the relationship between existing transfer-based attacks and the proposed method.

\subsection{Motivation}
Given a target classifier $f$ with parameters $\theta$ and 
a benign image $x \in \mathcal{X}$ where $x$ is in $d$ dimensions and $\mathcal{X}$ denotes all the legitimate images,
the adversarial attack aims to find an adversarial example $x^{adv} \in \mathcal{X}$ that satisfies:
\begin{equation}
    f(x;\theta) \neq f(x^{adv};\theta) \quad \text{ s.t. } \quad \|x - x^{adv}\| < \epsilon.
    \label{eq:attack_goal}
\end{equation}
For white-box attacks, we can regard the attack as an optimization problem that searches an example in the neighborhood of $x$ so as to maximize the loss function $J$ of the target classifier $f$:
\begin{equation}
    x^{adv} = \argmax_{\|x'-x\|_p<\epsilon} J(x',y;\theta).
\end{equation}

Lin \etal~\cite{lin2020nesterov} analogize the adversarial example generation process to the standard neural model training process, where the input $x$ can be viewed as parameters to be trained and the target model can be treated as the training set. From this perspective, the transferability of adversarial examples is equivalent to the generalization of the normally trained models. Therefore, existing works mainly focus on better optimization algorithms (\eg MI-FGSM, NI-FGSM) \cite{kurakin2017IFGSM, dong2018boosting, lin2020nesterov} or data augmentation (\eg ensemble attack on multiple models or input transformations) \cite{liu2017delving, xie2019improving, dong2019evading, lin2020nesterov, wang2021Admix} to improve the attack transferability.

In this work, we treat the iterative gradient-based adversarial attack as a stochastic gradient decent (SGD) optimization process, in which at each iteration, the attacker always chooses the target model for update.
As presented in previous works~\cite{roux2012stochastic, shalev2013stochastic, johnson2013accelerating}, SGD introduces large variance due to the randomness, leading to slow convergence. To address this issue, various variance reduction methods have been proposed to accelerate the convergence of SGD, \eg SAG (stochastic average gradient) \cite{roux2012stochastic}, SDCA (stochastic dual coordinate ascent) \cite{shalev2013stochastic}, and SVRG (stochastic variance reduced gradient) \cite{johnson2013accelerating}, which adopt the information from the training set to reduce the variance. 
Moreover, Nesterov's accelerated gradient \cite{Nesterov1983} that boosts the convergence, is beneficial to improve the attack transferability~\cite{lin2020nesterov}. 

Based on above analysis, we attempt to enhance the adversarial transferability with the gradient variance tuning strategy. The major difference between our method and SGD with variance reduction methods (SGDVRMs)~\cite{roux2012stochastic, shalev2013stochastic, johnson2013accelerating} is three-fold. First, we aim to craft highly transferable adversaries, which is equivalent to improving the generalization of the training models, while SGDVRMs aim to accelerate the convergence. Second, we consider the gradient variance of the examples sampled in the neighborhood of input $x$, which is equivalent to the one in the parameter space for training the neural models but SGDVRMs utilize variance in the training set. Third, our variance tuning strategy is more generalized and can be used to improve the performance of MI-FGSM and NI-FGSM. 

\begin{algorithm}[tb]
    \algnewcommand\algorithmicinput{\textbf{Input:}}
    \algnewcommand\Input{\item[\algorithmicinput]}
    \algnewcommand\algorithmicoutput{\textbf{Output:}}
    \algnewcommand\Output{\item[\algorithmicoutput]}

    \caption{VMI-FGSM}
    \label{alg:VMI-FGSM}
	\begin{algorithmic}[1]
		\Input A classifier $f$ with parameters $\theta$, loss function $J$
		\Input A raw example $x$ with ground-truth label $y$
		\Input The magnitude of perturbation $\epsilon$; number of iteration $T$ and decay factor $\mu$
		\Input The factor $\beta$ for the upper bound of neighborhood and number of example $N$ for variance tuning
        \Output An adversarial example $x^{adv}$
		\State $\alpha = \epsilon/T$
		\State $g_0 = 0$; $v_0=0$; $x_0^{adv}=x$
		\For{$t = 0 \rightarrow T-1$}
		    \State Calculate the gradient $\hat{g}_{t+1} = \bigtriangledown_{x^{adv}_t}J(x^{adv}_t, y; \theta)$
		    \State Update $g_{t+1}$ by variance tuning based momentum 
		    \begin{equation}
		        g_{t+1} = \mu \cdot g_t + \frac{\hat{g}_{t+1}+v_t}{\|\hat{g}_{t+1}+v_t\|_1}
		    \end{equation}
		    \State Update $v_{t+1} = V(x^{adv}_t)$ by Eq. \eqref{eq:variance}
		    \State Update $x_{t+1}^{adv}$ by applying the sign of gradient
		    \begin{equation}
		        x_{t+1}^{adv} = x_t^{adv} + \alpha \cdot \text{sign}(g_{t+1})
		    \end{equation}
		\EndFor
		\State $x^{adv}=x_T^{adv}$
        \State \Return $x^{adv}$
	\end{algorithmic} 
\end{algorithm} 

\subsection{Variance Tuning Gradient-based Attacks}
Typical gradient-based iterative attacks (\eg I-FGSM) greedily search an adversarial example in the direction of the sign of the gradient at each iteration, as shown in Eq.~\eqref{eq:ifgsm}, which may easily fall into poor local optima and ``overfit'' the model \cite{dong2018boosting}. MI-FGSM \cite{dong2018boosting} integrates momentum into I-FGSM for the purpose of stabilizing the update directions and escaping from poor local optima to improve the attack transferability. NI-FGSM \cite{lin2020nesterov} further adopts Netserov's accelerated gradient \cite{Nesterov1983} into I-FGSM to improve the transferability by leveraging its looking ahead property.

We observe that the above methods only consider the data points along the optimization path, denoted as $x_0^{adv}=x, x_1^{adv}, ..., x_{t-1}^{adv}, x_t^{adv}, ..., x_T^{adv} = x^{adv}$. 
In order to avoid overfitting and further improve the transferability of the adversarial attacks, we adopt the gradient information in the neighborhood of the previous data point to tune the gradient of the current data point at each iteration. 
Specifically, for any input $x \in \mathcal{X}$, we define the gradient variance as follows.
\begin{definition}
\textbf{Gradient Variance. } Given a classifier $f$ with parameters $\theta$ and loss function $J(x,y;\theta)$, an arbitrary image $x \in \mathcal{X}$ and an upper bound $\epsilon '$ for the neighborhood, the gradient variance can be defined as:
$$V_{\epsilon '}^g(x) = \mathbb{E}_{\|x' -x\|_p<\epsilon '} [\nabla_{x'} J(x', y; \theta)] \nonumber - \nabla_{x} J(x, y; \theta).$$
\end{definition}

We simply use $V(x)$ to denote $V_{\epsilon '}^g(x)$ without ambiguity in the following and we set $\epsilon ' = \beta \cdot \epsilon$ where $\beta$ is a hyper-parameter and $\epsilon$ is the upper bound of the perturbation magnitude.
In practice, however, due to the continuity of the input space, we cannot calculate $\mathbb{E}_{\|x' -x\|_p<\epsilon '} [\nabla_{x'} J(x', y; \theta)]$ directly. Therefore, we approximate its value by sampling $N$ examples in the neighborhood of $x$ to calculate $V(x)$:
\begin{equation}
    V(x) = \frac{1}{N}\sum_{i=1}^N \nabla_{x^i} J(x^i, y; \theta) - \nabla_{x} J(x, y; \theta). \label{eq:variance}
\end{equation}
Here $x^i = x + r_i, \  r_i \sim U[-(\beta \cdot \epsilon)^d, (\beta \cdot \epsilon)^d]$, and $U[a^d, b^d]$ stands for the uniform distribution in $d$ dimensions.

After obtaining the gradient variance, we can tune the gradient of $x_t^{adv}$ at the $t$-th iteration with the gradient variance $V(x_{t-1}^{adv})$ at the ($t-1$)-th iteration to stabilize the update direction. The algorithm of variance tuning MI-FGSM, denoted as VMI-FGSM, is summarized in Algorithm \ref{alg:VMI-FGSM}. Note that our method is generally applicable to any gradient-based attack method. We can easily extend VMI-FGSM to variance tuning NI-FGSM (VNI-FGSM), and integrate these methods with DIM, TIM and SIM as in \cite{lin2020nesterov}.

\begin{figure}
    \centering
    \includegraphics[width=\linewidth]{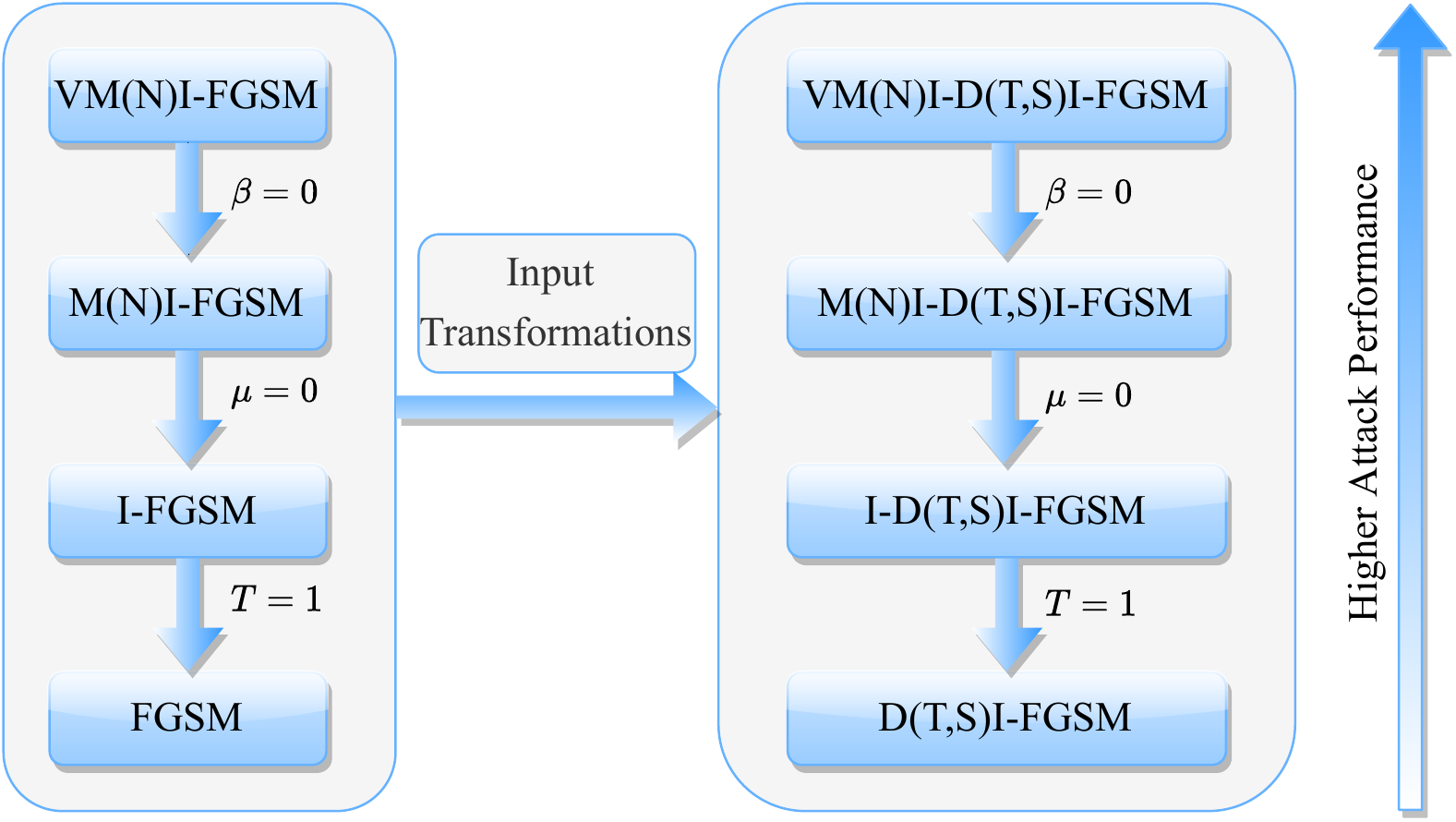}
    \caption{Relationships among various adversarial attacks. We can adjust the value of some hyper-parameters to relate the attacks derived from FSGM. We can also integrate various input transformations into these gradient-based attacks to further enhance the transferability. Here M(N)I-FGSM denotes MI-FGSM or NI-FGSM and D(T,S)I-FGSM denotes DI-FGSM, TI-FGSM or SI-FGSM.}
    \label{fig:relation}
\end{figure}

\begin{table*}[tb]
\small
\begin{center}
\begin{tabular}{|l|c|ccccccc|}
\hline
Model & Attack & Inc-v3 & Inc-v4 & IncRes-v2 & Res-101 & Inc-v3$_{ens3}$ & Inc-v3$_{ens4}$ & IncRes-v2$_{ens}$\\
\hline\hline
\multirow{4}{*}{Inc-v3} & MI-FGSM & \textbf{100.0*} & 43.6 & 42.4~~ & 35.7~~ & 13.1 & 12.8 &  ~~6.2\\
 & VMI-FGSM & \textbf{100.0*} & \textbf{71.7} & \textbf{68.1~~} & \textbf{60.2~~} & \textbf{32.8} & \textbf{31.2} & \textbf{17.5}\\
\cline{2-9}
 & NI-FGSM & \textbf{100.0*} & 51.7 & 50.3~~ & 41.3~~ & 13.5 & 13.2 &  ~~6.0\\
 & VNI-FGSM & \textbf{100.0*} & \textbf{76.5} & \textbf{74.9~~} & \textbf{66.0~~} & \textbf{35.0} & \textbf{32.8} & \textbf{18.8}\\
\hline
\multirow{4}{*}{Inc-v4} & MI-FGSM & 56.3 & ~~99.7* & 46.6~~ & 41.0~~ & 16.3 & 14.8 &  ~~7.5\\
 & VMI-FGSM & \textbf{77.9} & \textbf{~~99.8*} & \textbf{71.2~~} & \textbf{62.2~~} & \textbf{38.2} & \textbf{38.7} & \textbf{23.2}\\
\cline{2-9}
 & NI-FGSM & 63.1 & \textbf{100.0*} & 51.8~~ & 45.8~~ & 15.4 & 13.6 &  ~~6.7\\
 & VNI-FGSM & \textbf{83.4} & ~~99.9* & \textbf{76.1~~} & \textbf{66.9~~} & \textbf{40.0} & \textbf{37.7} & \textbf{24.5}\\
\hline
\multirow{4}{*}{IncRes-v2} & MI-FGSM & 60.7 & 51.1 & \textbf{97.9*} & 46.8~~ & 21.2 & 16.0 & 11.9\\
 & VMI-FGSM & \textbf{77.9} & \textbf{72.1} & \textbf{97.9*} & \textbf{67.7~~} & \textbf{46.4} & \textbf{40.8} & \textbf{34.4} \\
 \cline{2-9}
 & NI-FGSM & 62.8 & 54.7 & \textbf{99.1*} & 46.0~~ & 20.0 & 15.1 &  ~~9.6\\
 & VNI-FGSM & \textbf{80.8} & \textbf{76.9} & 98.5* & \textbf{69.8~~} & \textbf{47.9} & \textbf{40.3} & \textbf{34.2}\\
\hline
\multirow{4}{*}{Res-101} & MI-FGSM & 58.1 & 51.6 & 50.5~~ & \textbf{99.3*} & 23.9 & 21.5 & 12.7\\
 & VMI-FGSM & \textbf{75.1} & \textbf{68.9} & \textbf{70.5~~} & 99.2* & \textbf{45.2} & \textbf{41.4} & \textbf{30.1} \\
 \cline{2-9}
 & NI-FGSM & 65.6 & 58.3 & 57.0~~ & 99.4* & 24.5 & 21.4 & 11.7\\
 & VNI-FGSM & \textbf{79.8} & \textbf{74.6} & \textbf{73.2~~} & \textbf{99.7*} & \textbf{46.1} & \textbf{42.5} & \textbf{32.1}\\
\hline
\end{tabular}
\vspace{-0.5em}
\caption{The success rates (\%) on seven models in the single model setting by various gradient-based iterative attacks. The adversarial examples are crafted on Inc-v3, Inc-v4, IncRes-v2, and Res-101 respectively. * indicates the white-box model.}
\label{tab:singleModel}
\end{center}
\end{table*}

\subsection{Relationships among Various Attacks}
This work focuses on the transferability of adversarial attacks derived from FGSM. Here we summarize the relationships among these attacks, as illustrated in Figure \ref{fig:relation}. 
If the 
factor $\beta$ for the upper bound of neighborhood is set to $0$, VMI-FGSM and VNI-FGSM degrade to MI-FGSM and NI-FGSM, respectively. If the decay factor $\mu=0$, both MI-FGSM and NI-FGSM degrade to I-FGSM. If the iteration number $T=1$, I-FGSM degrades to FGSM. Moreover, we can integrate the above attacks with various input transformations, \ie DIM, TIM, SIM, to obtain more powerful adversarial attacks and these derived methods follow the same discipline.

\section{Experiments}
To validate the effectiveness of the proposed variance tuning based attack method, we conduct extensive experiments on standard ImageNet dataset \cite{russakovsky2015imagenet}. In this section, we first specify the experimental setup, then we compare our method with competitive baselines under various experimental settings and quantify the attack effectiveness on nine advanced defense models. Experimental results demonstrate that our method can significantly improve the transferability of the baselines in various settings. Finally, we provide further investigation on hyper-parameters $N$ and $\beta$ used for variance tuning.

\subsection{Experimental Setup}
\textbf{Dataset.} 
We randomly pick 1,000 clean images pertaining to the 1,000 categories from the ILSVRC 2012 validation set \cite{russakovsky2015imagenet}, which are almost correctly classified by all the testing models as in \cite{dong2018boosting, lin2020nesterov}.


\textbf{Models.} We consider four normally trained networks, including Inception-v3 (Inc-v3) \cite{szegedy2016inceptionv3}, Inception-v4 (Inc-v4), Inception-Resnet-v2 (IncRes-v2) \cite{szegedy2017inception} and Resnet-v2-101 (Res-101) \cite{he2016resnet} and three adversarially trained models, namely Inc-v3$_{ens3}$, Inc-v3$_{ens4}$ and IncRes-v2$_{ens}$ \cite{tramer2018ensemble}. Besides, we include nine advanced defense models that are robust against black-box adversarial attacks on the ImageNet dataset, \ie HGD \cite{liao2018defense}, R\&P \cite{xie2018mitigating}, NIPS-r3 \footnote{\url{https://github.com/anlthms/nips-2017/tree/master/mmd}}, Bit-Red \cite{xu2018BitReduction}, JPEG \cite{guo2018countering}, FD \cite{liu2019FD}, ComDefend \cite{jia2019comdefend}, RS \cite{cohen2019certified} and NRP \cite{naseer2020NRP}. For HGD, R\&P, NIPS-r3 and RS, we adopt the official models provided in corresponding papers. For all the other defense methods, we adopt Inc-v3$_{ens3}$ as the target model.

\textbf{Baselines.} We take two popular momentum based iterative adversarial attacks as our baselines, \ie MI-FGSM \cite{dong2018boosting} and NI-FGSM \cite{lin2020nesterov}, that exhibit better transferability than other white-box attacks \cite{goodfellow2015FGSM, kurakin2017IFGSM, carlini2017cw}. In addition, we integrate the proposed method with various input transformations, \ie DIM \cite{xie2019improving}, TIM \cite{dong2019evading}, SIM and CTM \cite{lin2020nesterov}, denoted as VM(N)I-DI-FGSM, VM(N)I-TI-FGSM, VM(N)I-SI-FGSM and VM(N)I-CT-FGSM respectively, to further validate the effectiveness of our method. 

\textbf{Hyper-parameters.} We follow the attack setting in  \cite{dong2018boosting} with the maximum perturbation of $\epsilon = 16$, number of iteration $T=10$ and step size $\alpha=1.6$. For MI-FGSM and NI-FGSM, we set the decay factor $\mu = 1.0$. For DIM, the transformation probability is set to $0.5$. For TIM, we adopt the Gaussian kernel with kernel size $7 \times 7$. For SIM, the number of scale copies is $5$ (\ie $i=0,1,2,3,4$). For the proposed method, we set $N=20$ and $\beta=1.5$.

\begin{figure*}
    \centering
    \begin{subfigure}{.33\textwidth}
        \centering 
        \includegraphics[width=\linewidth]{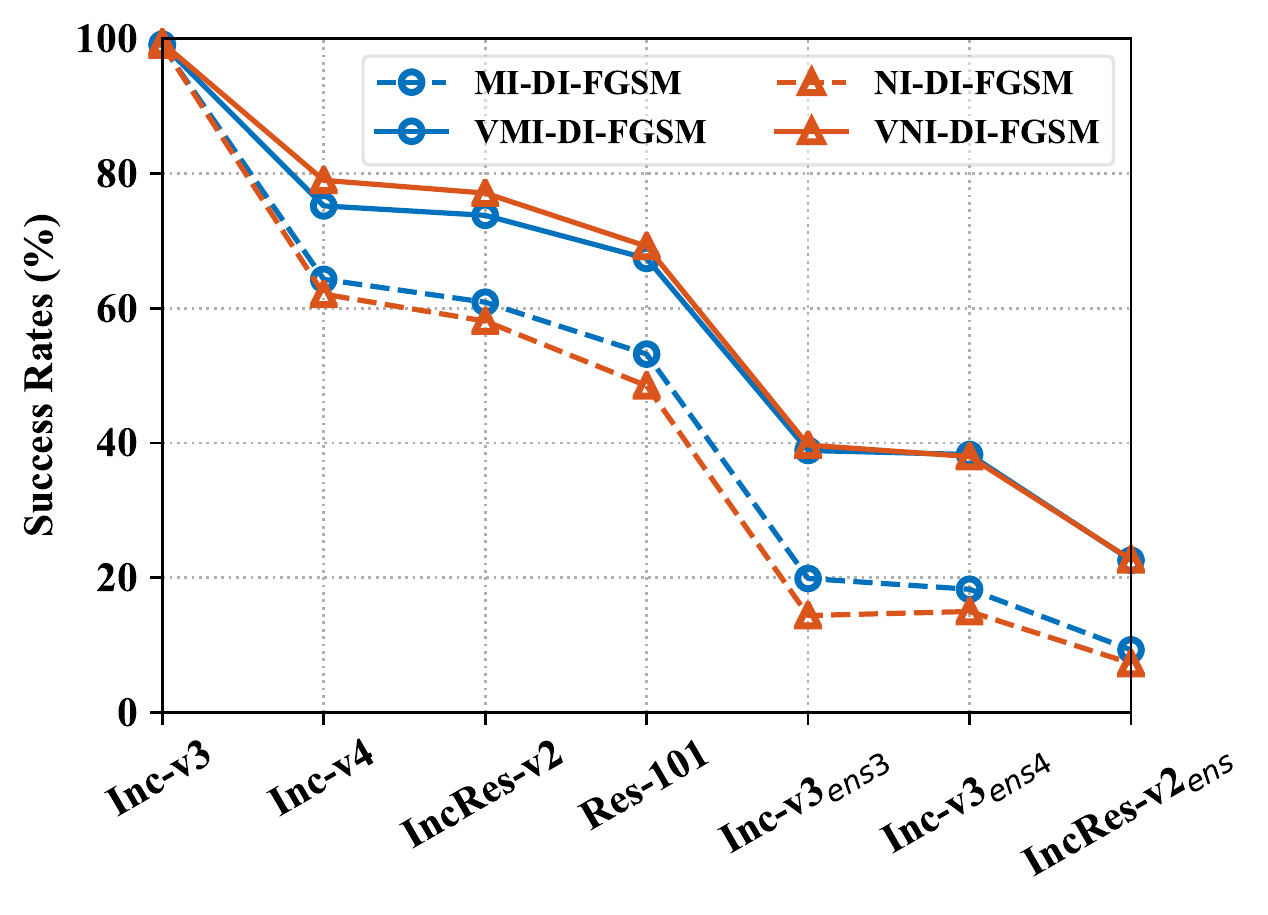}
        \vspace{-0.7cm}
        \caption{DIM}
    \end{subfigure}%
    \begin{subfigure}{.33\textwidth} 
        \centering 
        \includegraphics[width=\linewidth]{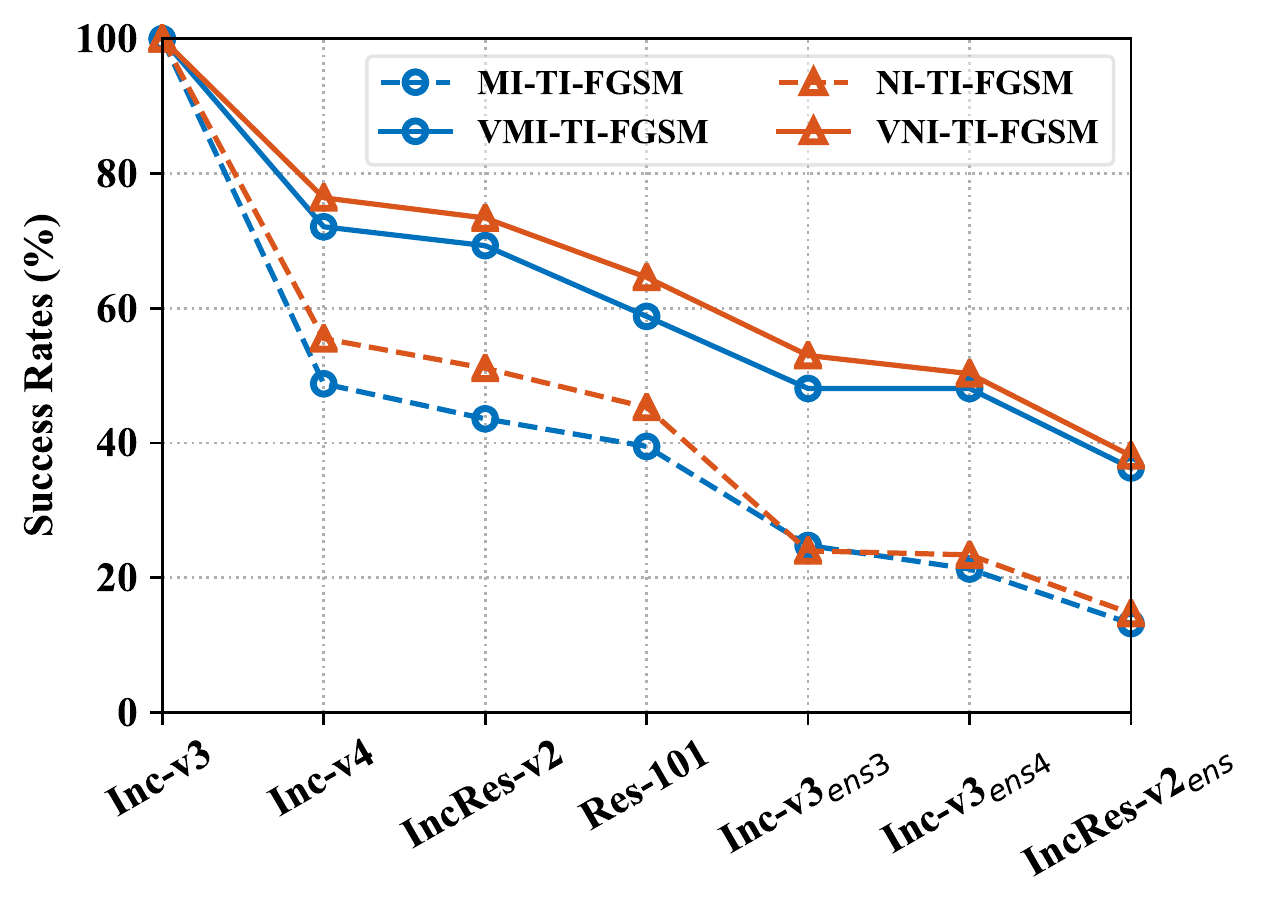}
        \vspace{-0.7cm}
        \caption{TIM}
    \end{subfigure}%
    \begin{subfigure}{.33\textwidth}
        \centering 
        \includegraphics[width=\linewidth]{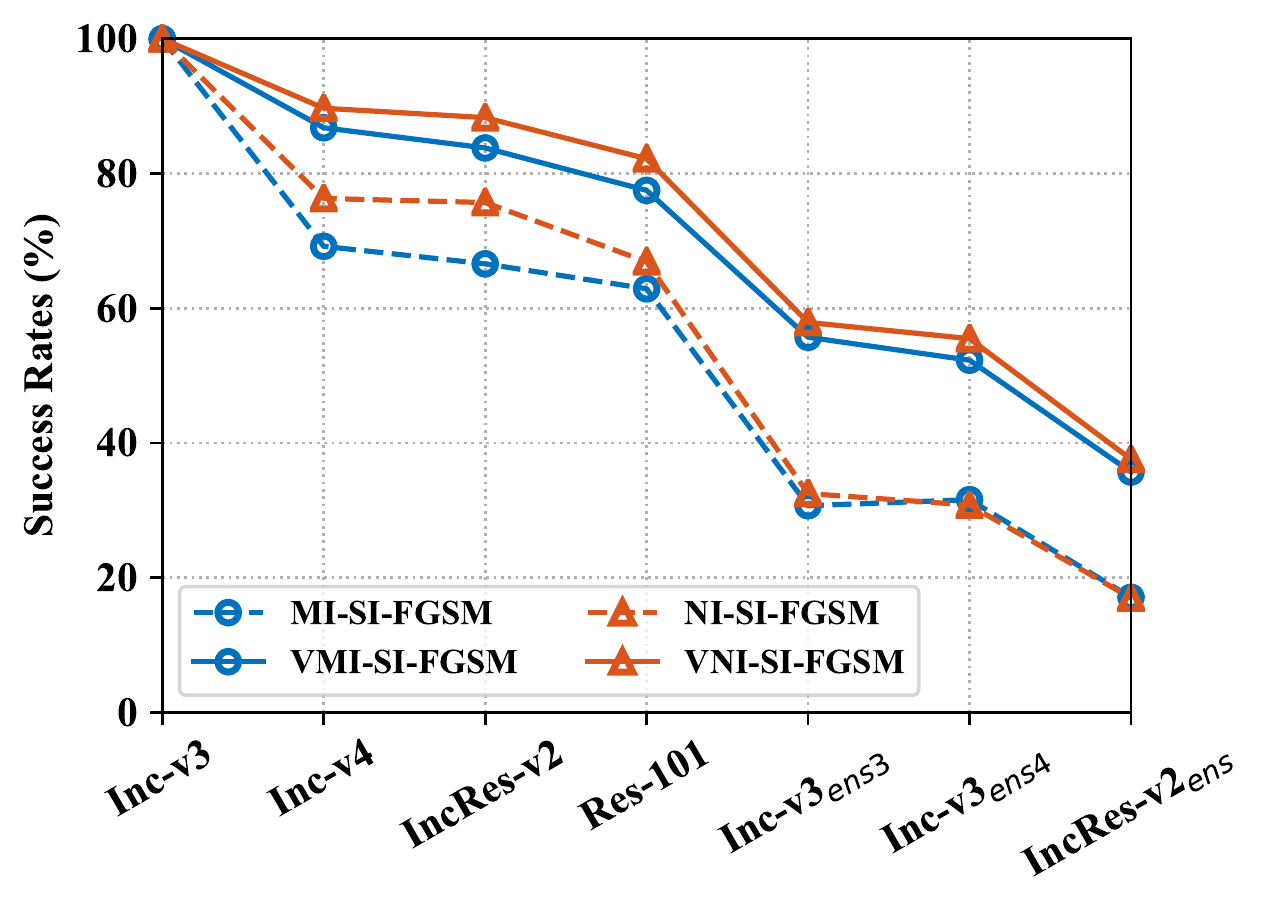}
        \vspace{-0.7cm}
        \caption{SIM}
    \end{subfigure}
    \vspace{-0.5em}
    \caption{The success rates (\%) on seven models in the single model setting by various gradient-based iterative attacks enhanced by DIM, TIM and SIM respectively. The adversarial examples are generated on Inc-v3 model.}
    \label{fig:transformation}
\end{figure*}
\begin{table*}[tb]
\small
\begin{center}
\begin{tabular}{|l|c|ccccccc|}
\hline
Model & Attack & Inc-v3 & Inc-v4 & IncRes-v2 & Res-101 & Inc-v3$_{ens3}$ & Inc-v3$_{ens4}$ & IncRes-v2$_{ens}$\\
\hline\hline
\multirow{4}{*}{Inc-v3} & MI-CT-FGSM & ~~98.7* & 85.4 & 80.6 & 76.0 & 64.1 & 62.1 & 45.2\\
 & VMI-CT-FGSM & \textbf{~~99.3*} & \textbf{88.6} & \textbf{86.7} & \textbf{82.9} & \textbf{78.6} & \textbf{76.2} & \textbf{64.7}\\
\cline{2-9}
 & NI-CT-FGSM & ~~98.9* & 84.1 & 80.0 & 74.5 & 60.0 & 56.2 & 41.0\\
 & VNI-CT-FGSM & \textbf{~~99.5*} & \textbf{91.2} & \textbf{89.0} & \textbf{85.3} & \textbf{78.6} & \textbf{76.7} & \textbf{65.3}\\
\hline
\multirow{4}{*}{Inc-v4} & MI-CT-FGSM & 87.2 & ~~98.6* & 83.3 & 78.3 & 72.2 & 67.2 & 57.3\\
 & VMI-CT-FGSM & \textbf{90.0} & \textbf{~~98.8*} & \textbf{86.6} & \textbf{81.9} & \textbf{78.3} & \textbf{76.6} & \textbf{68.3}\\
\cline{2-9}
 & NI-CT-FGSM & 87.8 & \textbf{~~99.4*} & 82.5 & 75.9 & 65.8 & 62.6 & 51.3\\
 & VNI-CT-FGSM & \textbf{92.1} & ~~99.2* & \textbf{89.2} & \textbf{85.1} & \textbf{80.1} & \textbf{78.3} & \textbf{70.4}\\
\hline
\multirow{4}{*}{IncRes-v2} & MI-CT-FGSM & 87.9 & 85.7 & \textbf{~~97.1*} & 83.0 & 77.6 & 74.6 & 72.0\\
 & VMI-CT-FGSM & \textbf{88.9} & \textbf{87.0} & ~~97.0* & \textbf{85.0} & \textbf{83.4} & \textbf{80.5} & \textbf{79.4}\\
 \cline{2-9}
 & NI-CT-FGSM & 90.2 & 87.0 & \textbf{~~99.4*} & 83.2 & 75.0 & 68.9 & 65.1\\
 & VNI-CT-FGSM & \textbf{92.9} & \textbf{90.6} & ~~99.0* & \textbf{88.2} & \textbf{85.2} & \textbf{82.5} & \textbf{81.8} \\
\hline
\multirow{4}{*}{Res-101} & MI-CT-FGSM & 86.5 & 81.8 & 83.2 & \textbf{~~98.9*} & 77.0 & 72.3 & 61.9\\
 & VMI-CT-FGSM & \textbf{86.9} & \textbf{84.2} & \textbf{86.4} & ~~98.6* & \textbf{81.0} & \textbf{78.6} & \textbf{71.6}\\
 \cline{2-9}
 & NI-CT-FGSM & 86.1 & 82.2 & 83.3 & ~~98.5* & 70.0 & 68.5 & 54.6\\
 & VNI-CT-FGSM &  \textbf{90.7} & \textbf{85.5} & \textbf{87.2} & \textbf{~~99.1*} & \textbf{82.6} & \textbf{79.7} & \textbf{73.3}\\
\hline
\end{tabular}
\vspace{-0.3em}
\caption{The success rates (\%) on seven models in the single model setting by various gradient-based iterative attacks enhanced by CTM. * indicates the white-box model.}
\label{tab:DI-TI-SIM}
\end{center}
\vspace{-1.0em}
\end{table*}
\begin{table*}[tb]
\small
\begin{center}
\begin{tabular}{|c|ccccccc|}
\hline
Attack & Inc-v3 & Inc-v4 & IncRes-v2 & Res-101 & Inc-v3$_{ens3}$ & Inc-v3$_{ens4}$ & IncRes-v2$_{ens}$\\
\hline\hline
MI-FGSM & \textbf{~~99.9*} & ~~98.2* & ~~95.3* & \textbf{~~99.9*} & 39.4 & 35.3 & 24.2\\
VMI-FGSM & ~~99.7* & \textbf{~~98.5*} & \textbf{~~96.0*} & \textbf{~~99.9*} & \textbf{67.6} & \textbf{62.9} & \textbf{50.7}\\
\hline
NI-FGSM & ~~99.8* & \textbf{~~99.8*} & \textbf{~~98.9*} & ~~99.8* & 41.0 & 33.5 & 23.1\\
VNI-FGSM & \textbf{~~99.9*} & ~~99.6* & ~~98.6* & \textbf{~~99.9*} & \textbf{71.3} & \textbf{66.0} & \textbf{52.9}\\
\hline
MI-CT-FGSM & ~~99.6* & ~~99.1* & ~~97.4* & ~~99.7* & 91.3 & 89.6 & 86.8\\
VMI-CT-FGSM & \textbf{~~99.7*} & \textbf{~~99.2*} & \textbf{~~98.4*} & \textbf{~~99.9*} & \textbf{93.6} & \textbf{92.4} & \textbf{91.0}\\
\hline
NI-CT-FGSM & \textbf{100.0*} & \textbf{100.0*} & \textbf{100.0*} & \textbf{100.0*} & 92.8 & 89.6 & 83.6\\
VNI-CT-FGSM &  \textbf{100.0*} & ~~99.9* & ~~99.6* & \textbf{100.0*} & \textbf{95.5} & \textbf{94.5} & \textbf{92.3}\\
\hline
\end{tabular}
\vspace{-0.3em}
\caption{The success rates (\%) on seven models in the multi-model setting by various gradient-based iterative attacks. The adversarial examples are generated on the ensemble models, \ie Inc-v3, Inc-v4, IncRes-v2 and Res-101.}
\label{tab:ensembleModels}
\end{center}
\vspace{-1.5em}
\end{table*}
\begin{table*}[tb]
\small
\begin{center}
\begin{tabular}{|c|c|cccccccccc|}
\hline
Model & Attack & HGD & R\&P & NIPS-r3 & Bit-Red & JPEG & FD & ComDefend & RS & NRP & Average\\
\hline\hline
\multirow{4}{*}{Inc-v3} & MI-CT-FGSM & 56.6 & 44.9 & 52.5 & 45.9 & 73.6 & 71.5 & 80.1 & 40.3 & 43.1 & 56.5 \\
& VMI-CT-FGSM & \textbf{73.1} & \textbf{65.1} & \textbf{70.3} & \textbf{59.1} & \textbf{82.2} & \textbf{78.4} & \textbf{86.0} & \textbf{51.9} & \textbf{60.2} & \textbf{69.6} \\
\cline{2-12}
& NI-CT-FGSM & 50.4 & 39.4 & 47.4 & 42.2 & 73.2 & 68.7 & 77.7 & 36.9 & 39.4 & 52.8 \\
& VNI-CT-FGSM & \textbf{73.4} & \textbf{64.5} & \textbf{70.6} & \textbf{57.9} & \textbf{85.0} & \textbf{80.5} & \textbf{87.3} & \textbf{52.1} & \textbf{58.9} & \textbf{70.0} \\
\hline
\multirow{4}{*}{Ens} & MI-CT-FGSM & 91.0 & 87.7 & 89.0 & 75.9 & 94.2 & 88.8 & 95.1 & 68.1 & 76.1 & 85.1 \\
& VMI-CT-FGSM & \textbf{92.9} & \textbf{91.0} & \textbf{92.3} & \textbf{80.9} & \textbf{95.4} & \textbf{91.0} & \textbf{96.2} & \textbf{77.0} & \textbf{83.2} & \textbf{88.9} \\
\cline{2-12}
& NI-CT-FGSM & 91.3 & 85.6 & 89.0 & 72.3 & 95.9 & 89.5 & 95.4 & 63.2 & 69.5 & 83.5 \\
& VNI-CT-FGSM & \textbf{94.7} & \textbf{92.4} & \textbf{93.4} & \textbf{82.3} & \textbf{97.1} & \textbf{92.6} & \textbf{97.4} & \textbf{77.4} & \textbf{84.0} & \textbf{90.1} \\
\hline
\end{tabular}
\vspace{-0.3em}
\caption{The success rates (\%) on nine models with advanced defense mechanism by various gradient-based iterative attacks enhanced by CTM. The adversarial examples are generated on Inc-v3 model and the ensemble of models respectively.}
\label{tab:defense}
\end{center}
\vspace{-1.5em}
\end{table*}

\subsection{Attack a Single Model}
We first perform four adversarial attacks, namely MI-FGSM, NI-FGSM, the proposed variance tuning based methods VMI-FGSM and VNI-FGSM, on a single neural network. We craft adversarial examples on normally trained networks and test them on all the seven neural networks we consider. The \textit{success rates}, which are the misclassification rates of the corresponding models on adversarial examples, are shown in Table \ref{tab:singleModel}. The models we attack are on rows and the seven models we test are on columns.

We can observe that VMI-FGSM and VNI-FGSM outperform the baseline attacks by a large margin on all the black-box models, while maintain high success rates on all the white-box models. For instance, if we craft adversarial examples on Inc-v3 model in which all the attacks can achieve $100\%$ success rates in the white-box setting, VMI-FGSM yields $71.7\%$ success rate on Inc-v4 and $32.8\%$ success rate on Inc-v3$_{ens3}$, while the baseline MI-FGSM only obtains the corresponding success rates of $43.6\%$ and $13.1\%$, respectively. This convincingly validates the high effectiveness of the proposed method. We also illustrate several adversarial images generated on Inc-v3 model by various attack methods in Appendix \ref{app:sec:adv}, showing that these generated adversarial perturbations are all human imperceptible but our method leads to higher transferability.

\subsection{Attack with Input Transformations}
Several input transformations, \eg DIM, \cite{dong2019evading}, TIM \cite{xie2019improving} and SIM \cite{lin2020nesterov}, have been incorporated into the gradient-based adversarial attacks, which are effective to improve the transferability. Here we integrate our methods into these input transformations and demonstrate the proposed variance tuning strategy could further enhance the transferability. We report the success rates of black-box attacks in Figure  \ref{fig:transformation}, where the adversarial examples are generated on Inc-v3 model. The results for adversarial examples generated on other three models are reported in Appendix \ref{app:sec:attack}.

The results show that the success rates against black-box models are improved by a large margin with the variance tuning strategy regardless of the attack algorithms or the white-box models to be attacked. In general, the methods equipped with our variance tuning strategy consistently outperform the baseline attacks by $10\% \sim 30\%$. 

Lin \etal \cite{lin2020nesterov} have shown that CTM, the combination of DIM, TIM and SIM, could help the gradient-based attacks achieve 
great 
transferability. We also combine CTM with our method to further improve the transferability. As depicted in Table \ref{tab:DI-TI-SIM}, the success rates could be further improved remarkably on various models, especially against adversarially trained models, which further demonstrates the high effectiveness and generalization of 
our method.

\subsection{Attack an Ensemble of Models}
\label{sec:ensembel}
Liu \etal \cite{liu2017delving} have shown that attacking multiple models simultaneously could improve the transferability of the generated adversarial examples. In this subsection, we adopt the ensemble attack method in \cite{dong2018boosting}, which fuses the logit outputs of different models, to demonstrate that our variance tuning method could further improve the transferability of adversarial attacks in the multi-model setting. Specifically, we attack the ensemble of four normally trained models, \ie Inc-v3, Inc-v4, IncRes-v2 and Res-101 by averaging the logit outputs of the models using various attacks with or without input transformations. 

As shown in Table \ref{tab:ensembleModels}, our methods (VMI-FGSM, VNI-FGSM) could significantly enhance the transferability of the baselines more than $25\%$ for MI-FGSM and $30\%$ for NI-FGSM on the adversarially trained models. Even though the attacks with CTM could achieve good enough transferability, our methods (VMI-CT-FGSM, VNI-CT-FGSM) could further improve the transferability significantly. In particular, VNI-CT-FGSM achieves the success rates of $92.3\% \sim 95.5\%$ against three adversarially trained models, indicating the vulnerability of current defense mechanisms. Besides, in the white-box setting, our methods could still maintain similar success rates as the baselines. 

\begin{figure*}
\centering 
    \begin{minipage}[b]{0.48\textwidth} 
        \begin{subfigure}{.48\textwidth}
          \centering 
          \includegraphics[width=\linewidth]{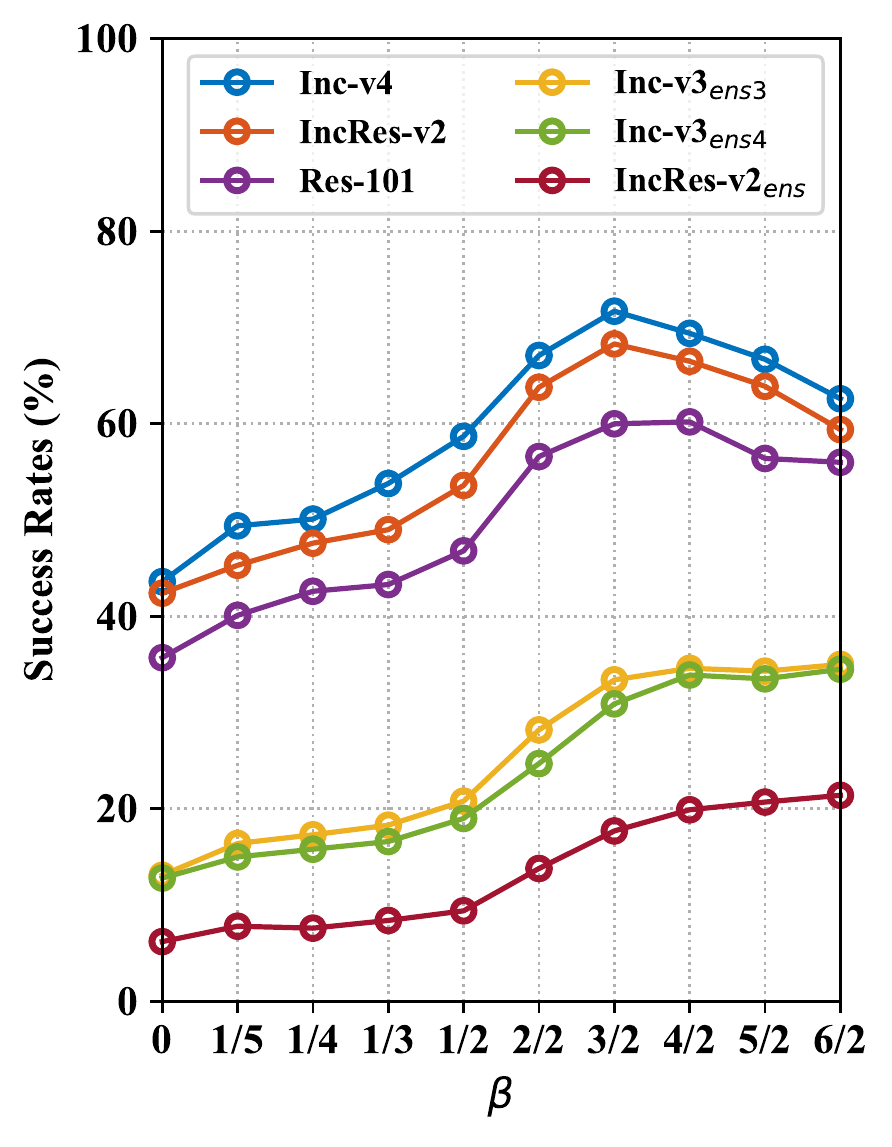}
          \caption{VMI-FGSM}
          \label{fig:beta:vmi}
        \end{subfigure}
        \hspace{.1em}
        \begin{subfigure}{.48\textwidth} 
          \centering 
          \includegraphics[width=\linewidth]{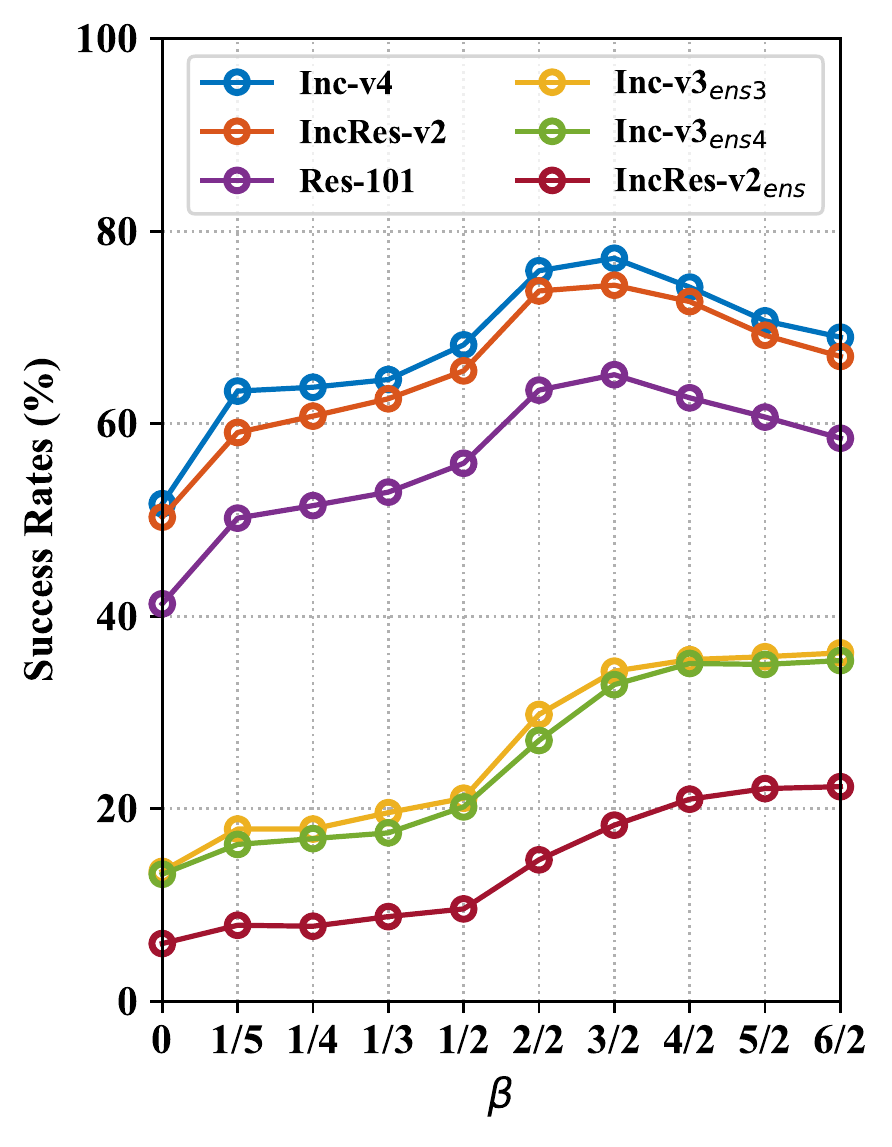}
          \caption{VNI-FGSM}
          \label{fig:beta:vni}
        \end{subfigure}%
    \vspace{-0.3em}
    \caption{The success rates (\%) on the other six models with adversarial examples generated by VMI-FGSM and VNI-FGSM on Inc-v3 model when varying factor $\beta$ for the upper bound of the neighborhood.}
    \label{fig:beta}
    \end{minipage}
    \hspace{.3cm}
    \begin{minipage}[b]{0.48\textwidth} 
        \begin{subfigure}{.48\textwidth}
          \centering 
          \includegraphics[width=\linewidth]{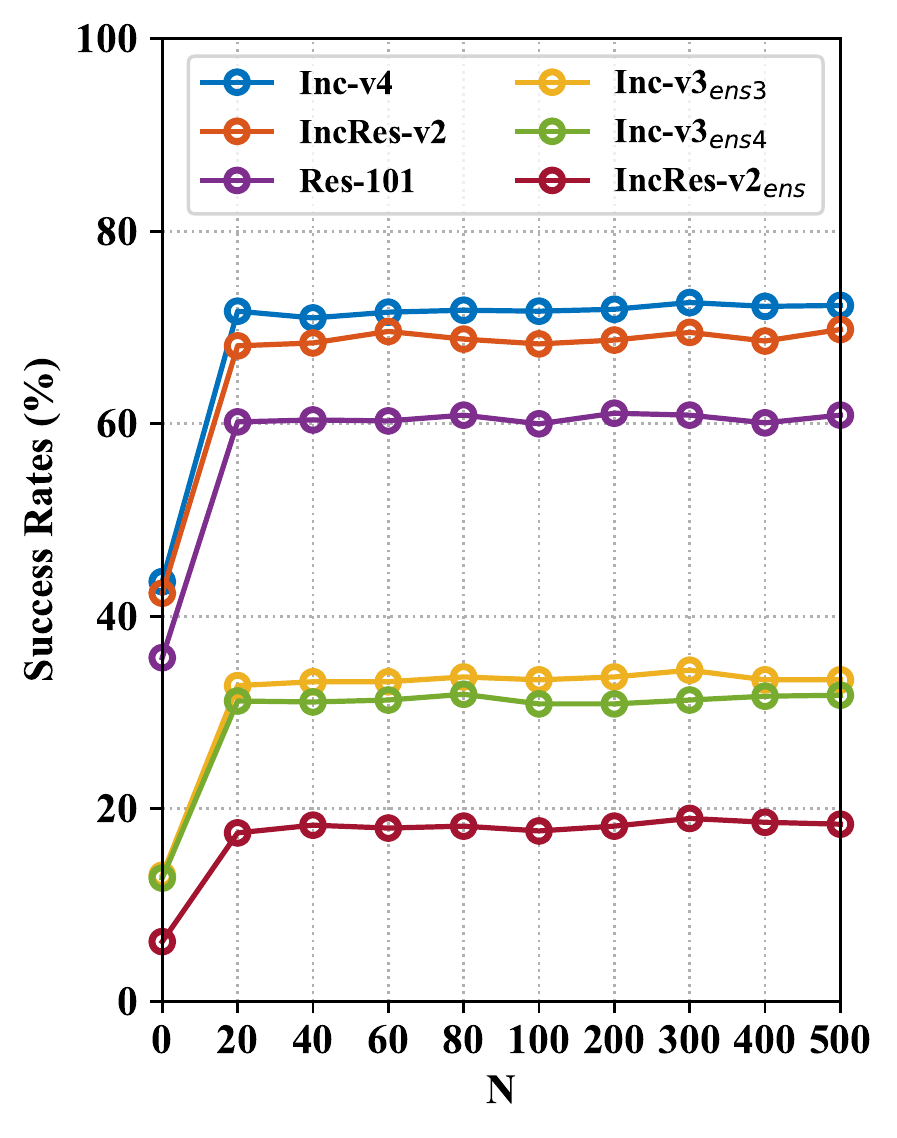}
          \caption{VMI-FGSM}
          \label{fig:number:vmi}
        \end{subfigure}
        \hspace{.1em}
        \begin{subfigure}{.48\textwidth} 
          \centering 
          \includegraphics[width=\linewidth]{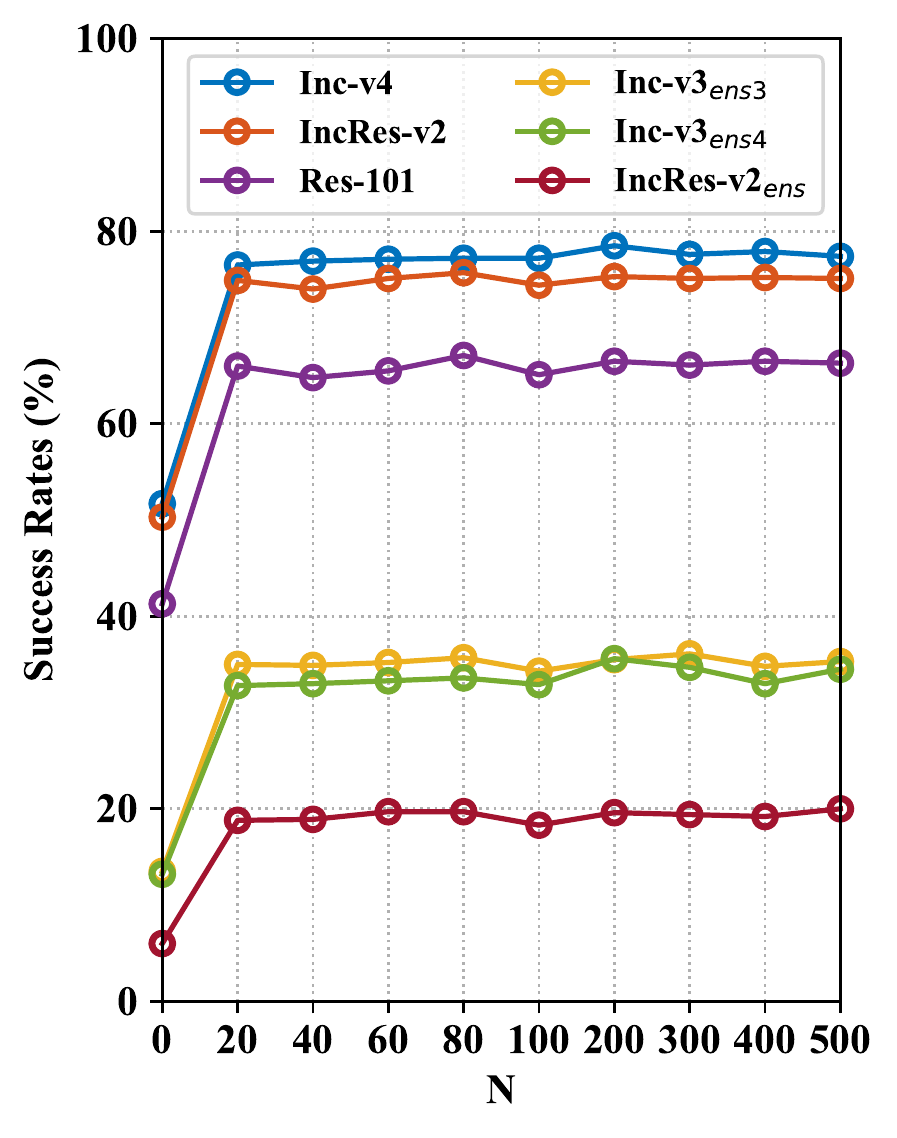}
          \caption{VNI-FGSM}
          \label{fig:number:vni}
        \end{subfigure}%
    \vspace{-0.3em}
    \caption{The success rates (\%) on the other six models with adversarial examples generated by VMI-FGSM and VNI-FGSM on Inc-v3 model when varying the number of sampled example $N$.}
    \label{fig:number}
    \end{minipage}
    \vspace{-0.5em}
\end{figure*}
\subsection{Attack Advanced Defense Models}
To further validate the effectiveness of the proposed method in practice, except for the normally trained models and adversarially trained models, we also evaluate our methods on nine extra models with advanced defenses, including the top-3 defense methods in the NIPS competition (HGD (rank-1) \cite{liao2018defense}, R\&P (rank-2) \cite{xie2018mitigating} and NIPS-r3 (rank-3)), and six recently proposed defense methods, namely Bit-Red \cite{xu2018BitReduction}, JPEG \cite{guo2018countering}, FD \cite{liu2019FD}, ComDefend \cite{jia2019comdefend}, RS \cite{cohen2019certified} and NRP \cite{naseer2020NRP}.

Since MI-CT-FGSM and NI-CT-FGSM exhibit the best transferability among the existing attack methods~\cite{lin2020nesterov}, we compare our methods with the two attacks with adversarial examples crafted on Inc-v3 model and the ensemble models as in Section \ref{sec:ensembel}, respectively. The results are shown in Table \ref{tab:defense}. In the single model setting, the proposed methods achieve an average success rate of $66.5\%$ for VMI-CT-FGSM and $67.0\%$ for VNI-CT-FGSM, which outperforms the baseline attacks for more than $13.5\%$ and $17.5\%$ respectively. In the multi-model setting, our methods achieve an average success rate of $88.9\%$ for VMI-CT-FGSM and $90.1\%$ for VNI-CT-FGSM, which outperforms the baseline attacks for more than $3.8\%$ and $6.6\%$ respectively. Note that in the multi-model setting, our methods achieve the success rates of more than $77\%$ against the defense model with Randomize Smoothing (RS) \cite{cohen2019certified} that provides certified defense. And our methods achieve the success rates of more than $83\%$ against the defense model with Neural Representation Purifier (NRP) which is a recently proposed powerful defense method and exhibits great robustness against DIM and DI-TIM \cite{naseer2020NRP}, raising a new security issue for the development of more robust deep learning models.

\subsection{Ablation Study on Hyper-parameters}
We conduct a series of ablation experiments to study the impact of two hyper-parameters of the proposed variance tuning, $N$ and $\beta$. All the adversarial examples are generated on Inc-v3 model without input transformations, which achieve success rates of $100\%$ for different values of the two parameters in the white-box setting.

\textbf{On the upper bound of neighborhood.} In Figure \ref{fig:beta}, we study the influence of the neighborhood size, determined by parameter $\beta$, on the success rates in the black-box setting where $N$ is fixed to $20$. When $\beta = 0$, VMI-FGSM and VNI-FGSM degrade to MI-FGSM and NI-FGSM, respectively, and achieve the lowest transferability. When $\beta = 1/5$, although the neighborhood is very small, our variance tuning strategy could improve the transferability remarkably. As we increase $\beta$, the transferability increases and achieves the peak for normally trained models when $\beta = 3/2$ but it is still increasing against the adversarially trained models. In order to achieve the trade-off for the transferability on normally trained models and adversarially trained models, we choose $\beta = 3/2$ in our experiments.

\textbf{On the number of sampled examples in the neighborhood.} We then study the influence of the number of sampled examples $N$ on the success rates in the black-box setting ($\beta$ is fixed to $3/2$). As depicted in Figure \ref{fig:number}, when $N=0$, VMI-FGSM and VNI-FGSM also degrade to MI-FGSM and NI-FGSM, respectively, and achieve the lowest transferability. When $N=20$, the transferability of adversarial examples is improved significantly and the transferability increases slowly when we continually increase $N$. Note that we need $N$ forward and backward propogations for gradient variance at each iteration as shown in Eq.~\ref{eq:variance}, thus a bigger value of $N$ means a higher computation cost. To balance the transferability and computation cost, we set $N=20$ in our experiments.

In summary, $\beta$ plays a key role in improving the transferability while $N$ exhibits little impact when $N>20$. In our experiments, we set $\beta=3/2$ and $N=20$.

\section{Conclusion}
In this work, we propose a variance tuning method to enhance the transferability of the iterative gradient-based adversarial attacks. Specifically, for any input of a neural classifier, we define its gradient variance as the difference between the mean gradient of the neighborhood and its own gradient. Then we adopt the gradient variance of the data point at the previous iteration along the optimization path to tune the current gradient.
Extensive experiments demonstrate that the variance tuning method could significantly improve the transferability of the existing competitive attacks, MI-FGSM and NI-FGSM, while maintain similar success rates in the white-box setting. 

The variance tuning method is generally applicable to any iterative gradient-based attacks.
We employ our method to attack ensemble models and then integrate with advanced input transformation methods (\eg DIM, TIM, SIM) to further enhance the transferability. Empirical results on nine advanced defense models show that our integrated method could reach an average success rate of at least $90.1\%$, outperforming the state-of-the-art attacks for $6.6\%$ on average, indicating the insufficiency of current defense techniques.


\section*{Acknowledgements}

This work is supported by National Natural Science Foundation (62076105) and Microsft Research Asia Collaborative Research Fund (99245180).

{\small
\bibliographystyle{ieee_fullname}
\bibliography{egbib}
}

\newpage
~~
\newpage
\appendix
{\centering
\section*{Appendix}}
In the appendix, we first visualize more adversarial examples generated by various attacks. Then we provide more results for our methods integrated with DIM, TIM or SIM on the other three normally trained models, \ie Inc-v4, IncRes-v2 and Res-101.

\section{Visualizations on Adversarial Examples}
\label{app:sec:adv}
\begin{figure*}[b]
    \vspace{-5em}
    \centering
    \begin{minipage}[c]{0.115\textwidth}\raggedleft
        \begin{subfigure}{\textwidth}
          \centering 
          \makebox[0pt][r]{\makebox[20pt]{\raisebox{25pt}{\rotatebox[origin=c]{90}{Raw Image}}}}%
          \includegraphics[width=\linewidth]{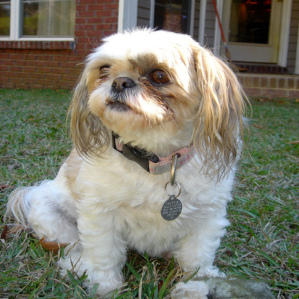} \\
          \vspace{0.3em}
          \makebox[0pt][r]{\makebox[20pt]{\raisebox{23pt}{\rotatebox[origin=c]{90}{MI-FGSM}}}}%
          \includegraphics[width=\linewidth]{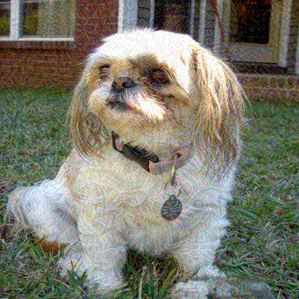} \\
          \vspace{0.3em}
          \makebox[0pt][r]{\makebox[20pt]{\raisebox{23pt}{\rotatebox[origin=c]{90}{NI-FGSM}}}}%
          \includegraphics[width=\linewidth]{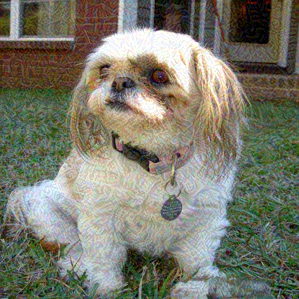} \\
          \vspace{0.3em}
          \makebox[0pt][r]{\makebox[20pt]{\raisebox{23pt}{\rotatebox[origin=c]{90}{VMI-FGSM}}}}%
          \includegraphics[width=\linewidth]{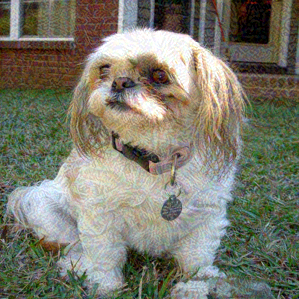} \\
          \vspace{0.3em}
          \makebox[0pt][r]{\makebox[20pt]{\raisebox{23pt}{\rotatebox[origin=c]{90}{VNI-FGSM}}}}%
          \includegraphics[width=\linewidth]{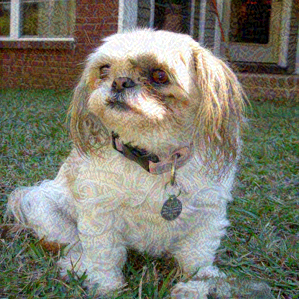} 
        \end{subfigure}
    \end{minipage}
    \hspace{-0.05cm}
    \begin{minipage}[c]{0.115\textwidth} 
        \begin{subfigure}{\textwidth}
          \centering 
          \includegraphics[width=\linewidth]{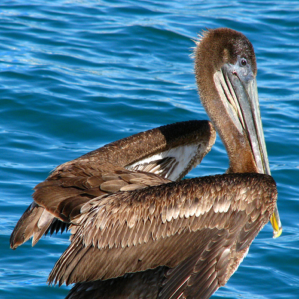}\\
          \vspace{0.3em}
          \includegraphics[width=\linewidth]{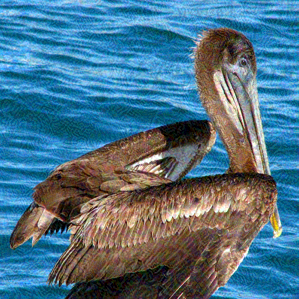}\\
          \vspace{0.3em}
          \includegraphics[width=\linewidth]{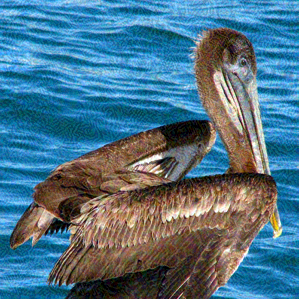}\\
          \vspace{0.3em}
          \includegraphics[width=\linewidth]{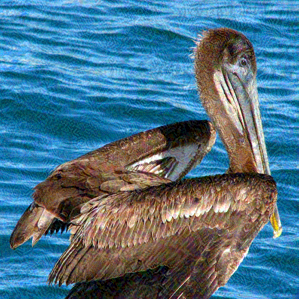}\\
          \vspace{0.3em}
          \includegraphics[width=\linewidth]{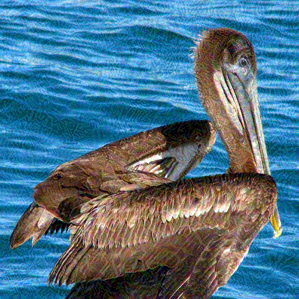}
        \end{subfigure}
    \end{minipage}
     \hspace{-0.05cm}
    \begin{minipage}[c]{0.115\textwidth} 
        \begin{subfigure}{\textwidth}
          \centering 
          \includegraphics[width=\linewidth]{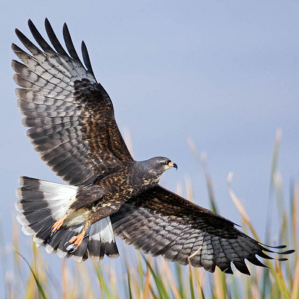}\\
          \vspace{0.3em}
          \includegraphics[width=\linewidth]{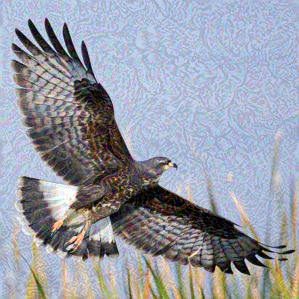}\\
          \vspace{0.3em}
          \includegraphics[width=\linewidth]{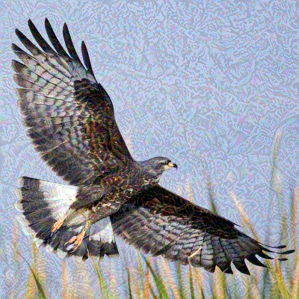}\\
          \vspace{0.3em}
          \includegraphics[width=\linewidth]{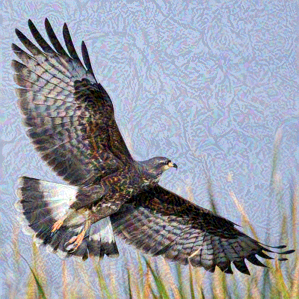}\\
          \vspace{0.3em}
          \includegraphics[width=\linewidth]{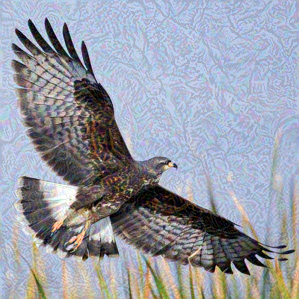}
        \end{subfigure}
    \end{minipage}
      \hspace{-0.05cm}
    \begin{minipage}[c]{0.115\textwidth} 
        \begin{subfigure}{\textwidth}
          \centering 
          \includegraphics[width=\linewidth]{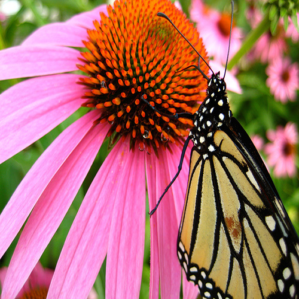}\\
          \vspace{0.3em}
          \includegraphics[width=\linewidth]{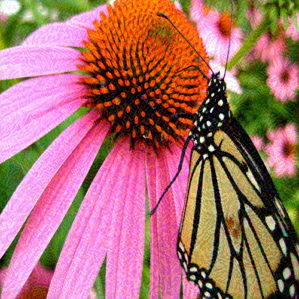}\\
          \vspace{0.3em}
          \includegraphics[width=\linewidth]{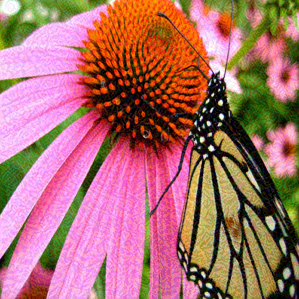}\\
          \vspace{0.3em}
          \includegraphics[width=\linewidth]{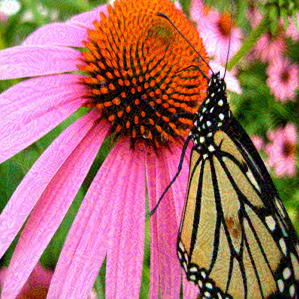}\\
          \vspace{0.3em}
          \includegraphics[width=\linewidth]{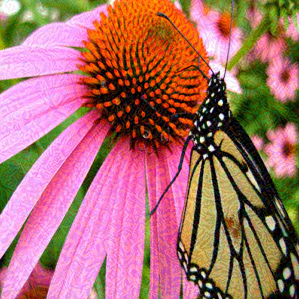}
        \end{subfigure}
    \end{minipage}
      \hspace{-0.05cm}
    \begin{minipage}[c]{0.115\textwidth} 
        \begin{subfigure}{\textwidth}
          \centering 
          \includegraphics[width=\linewidth]{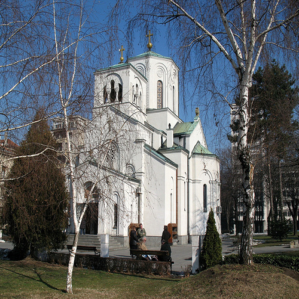}\\
          \vspace{0.3em}
          \includegraphics[width=\linewidth]{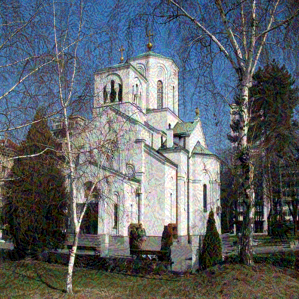}\\
          \vspace{0.3em}
          \includegraphics[width=\linewidth]{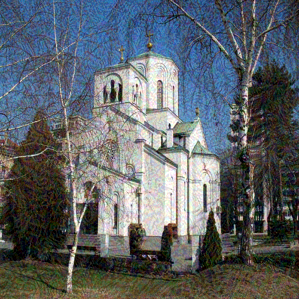}\\
          \vspace{0.3em}
          \includegraphics[width=\linewidth]{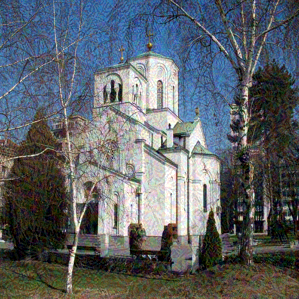}\\
          \vspace{0.3em}
          \includegraphics[width=\linewidth]{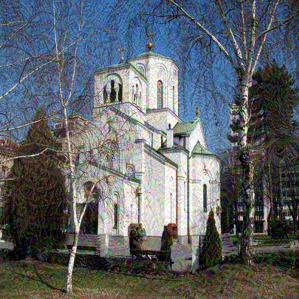}
        \end{subfigure}
    \end{minipage}
      \hspace{-0.05cm}
    \begin{minipage}[c]{0.115\textwidth} 
        \begin{subfigure}{\textwidth}
          \centering 
          \includegraphics[width=\linewidth]{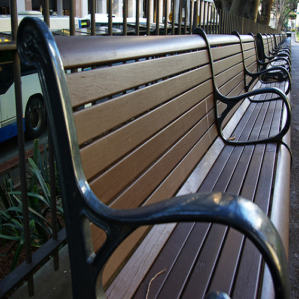}\\
          \vspace{0.3em}
          \includegraphics[width=\linewidth]{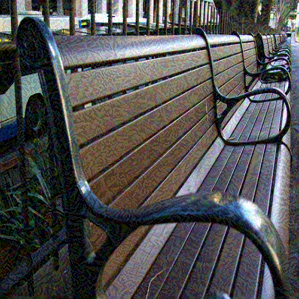}\\
          \vspace{0.3em}
          \includegraphics[width=\linewidth]{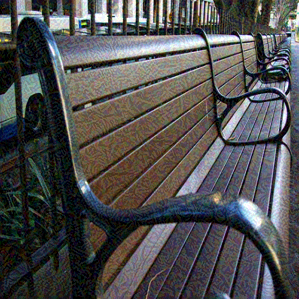}\\
          \vspace{0.3em}
          \includegraphics[width=\linewidth]{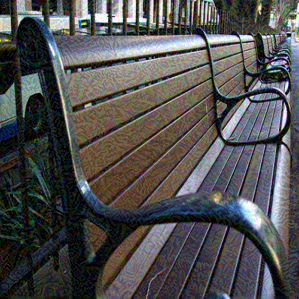}\\
          \vspace{0.3em}
          \includegraphics[width=\linewidth]{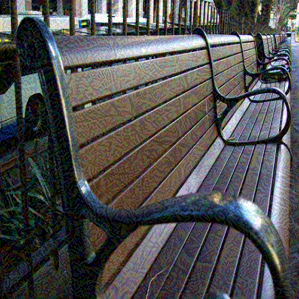}
        \end{subfigure}
    \end{minipage}
      \hspace{-0.05cm}
    \begin{minipage}[c]{0.115\textwidth} 
        \begin{subfigure}{\textwidth}
          \centering 
          \includegraphics[width=\linewidth]{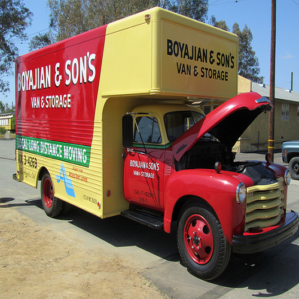}\\
          \vspace{0.3em}
          \includegraphics[width=\linewidth]{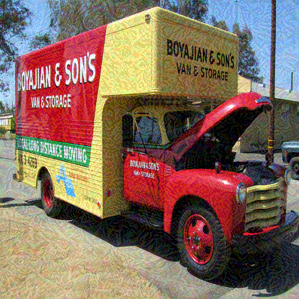}\\
          \vspace{0.3em}
          \includegraphics[width=\linewidth]{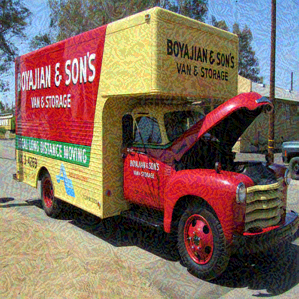}\\
          \vspace{0.3em}
          \includegraphics[width=\linewidth]{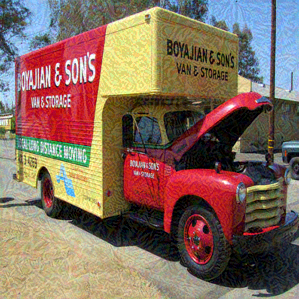}\\
          \vspace{0.3em}
          \includegraphics[width=\linewidth]{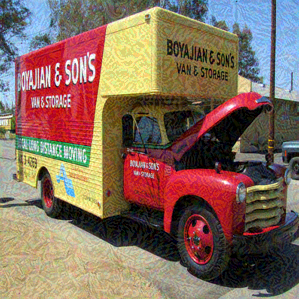}
        \end{subfigure}
    \end{minipage}
      \hspace{-0.05cm}
    \begin{minipage}[c]{0.115\textwidth} 
        \begin{subfigure}{\textwidth}
          \centering 
          \includegraphics[width=\linewidth]{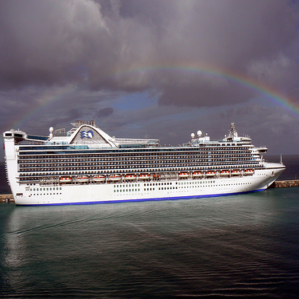}\\
          \vspace{0.3em}
          \includegraphics[width=\linewidth]{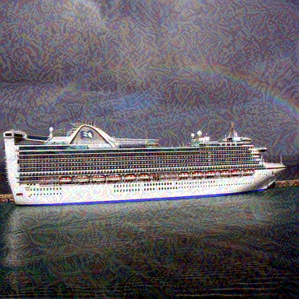}\\
          \vspace{0.3em}
          \includegraphics[width=\linewidth]{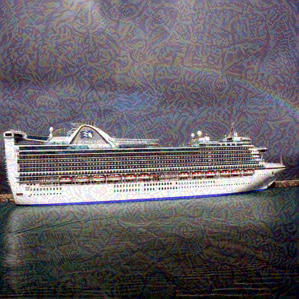}\\
          \vspace{0.3em}
          \includegraphics[width=\linewidth]{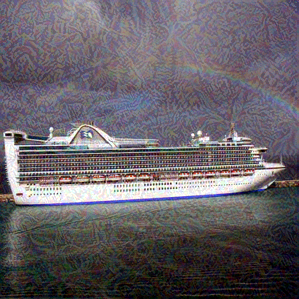}\\
          \vspace{0.3em}
          \includegraphics[width=\linewidth]{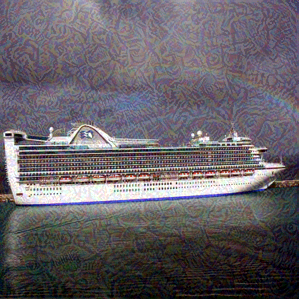}
        \end{subfigure}
    \end{minipage}
    \caption{Adversarial examples generated by MI-FGSM \cite{dong2018boosting}, NI-FGSM \cite{lin2020nesterov}, the proposed VMI-FGSM and VNI-FGSM on Inc-v3 model \cite{szegedy2016inceptionv3} with the maximum perturbation of $\epsilon = 16$.}
    \label{app:fig:adv_images}
\end{figure*}

We visualize eight randomly selected benign images and their corresponding adversarial examples crafted by various attacks in Figure  \ref{app:fig:adv_images}. The adversarial examples are crafted on Inc-v3 model, using MI-FGSM, NI-FGSM, VMI-FGSM and VNI-FGSM, respectively. It can be observed that these crafted adversarial examples are human-imperceptible.

\section{More Attacks with Input Transformations}
\label{app:sec:attack}
\begin{figure*}
    \centering
    \begin{subfigure}{.33\textwidth}
          \centering 
          \includegraphics[width=\linewidth]{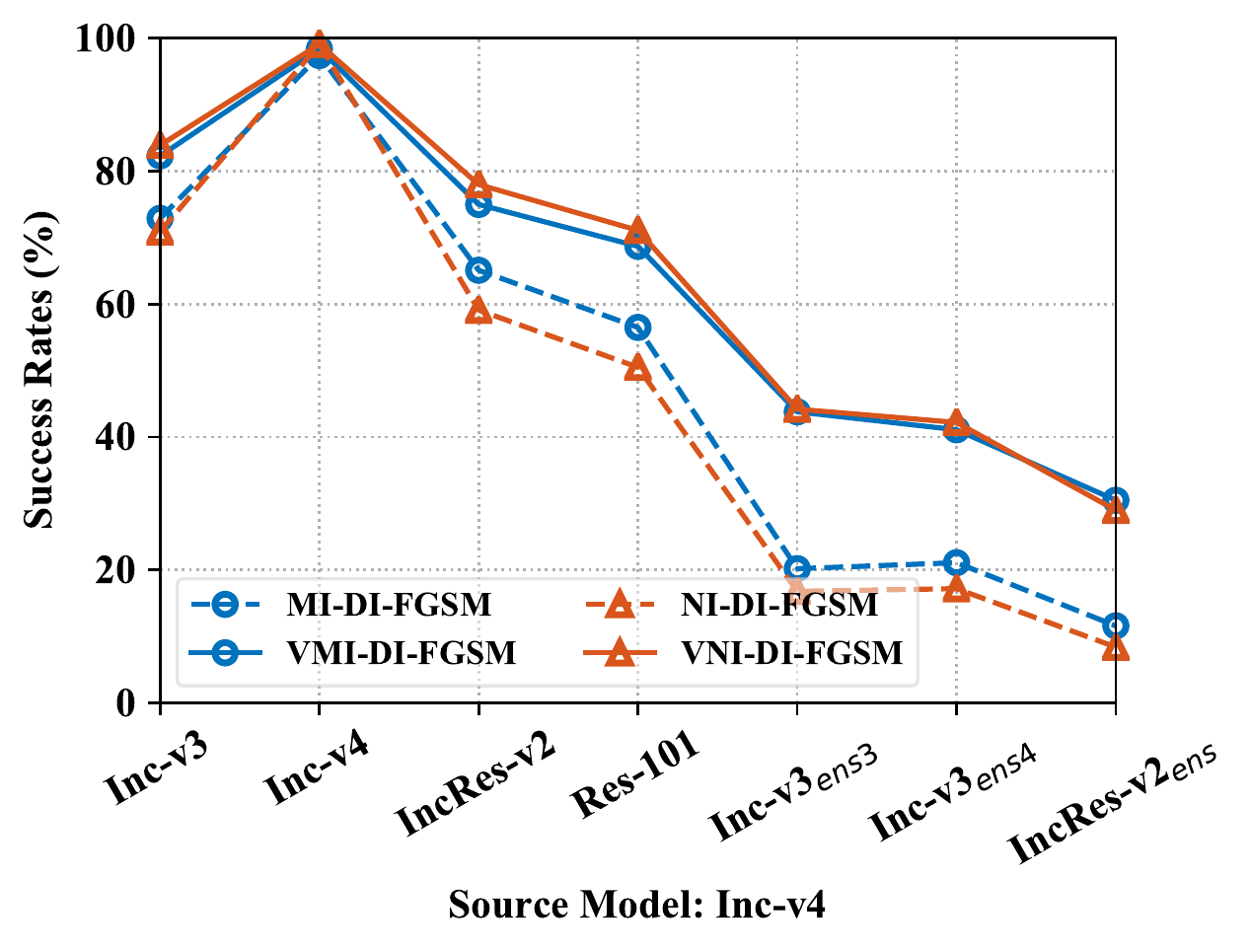}
        \end{subfigure}
        \begin{subfigure}{.33\textwidth} 
          \centering 
          \includegraphics[width=\linewidth]{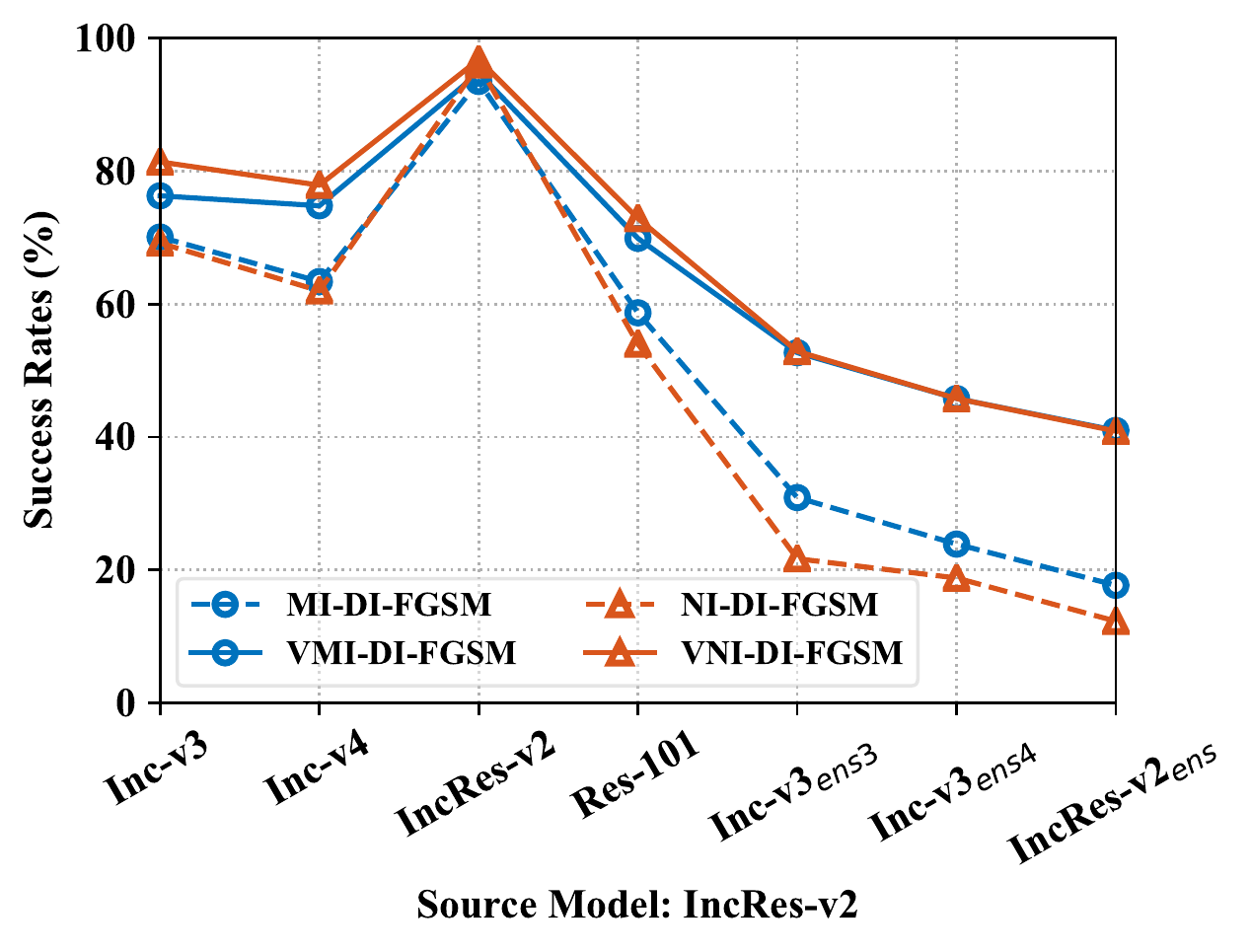}
        \end{subfigure}%
        \begin{subfigure}{.33\textwidth}
          \centering 
          \includegraphics[width=\linewidth]{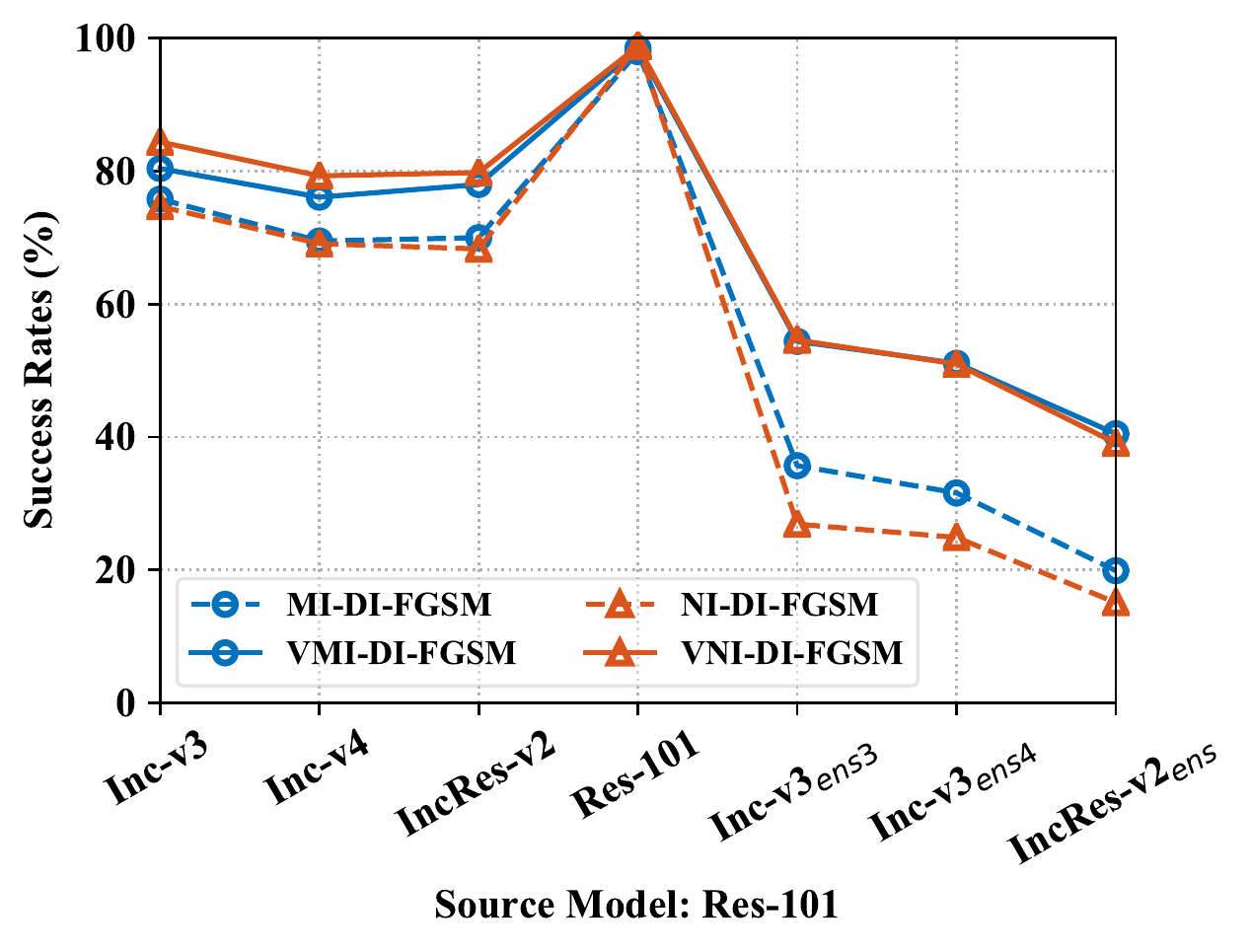}
        \end{subfigure}
    \caption{The success rates (\%) on seven models in the single model setting by various gradient-based iterative attacks enhanced by DIM.}
    \label{app:fig:dim}
    \vspace{1em}
    \centering
    \begin{subfigure}{.33\textwidth}
          \centering 
          \includegraphics[width=\linewidth]{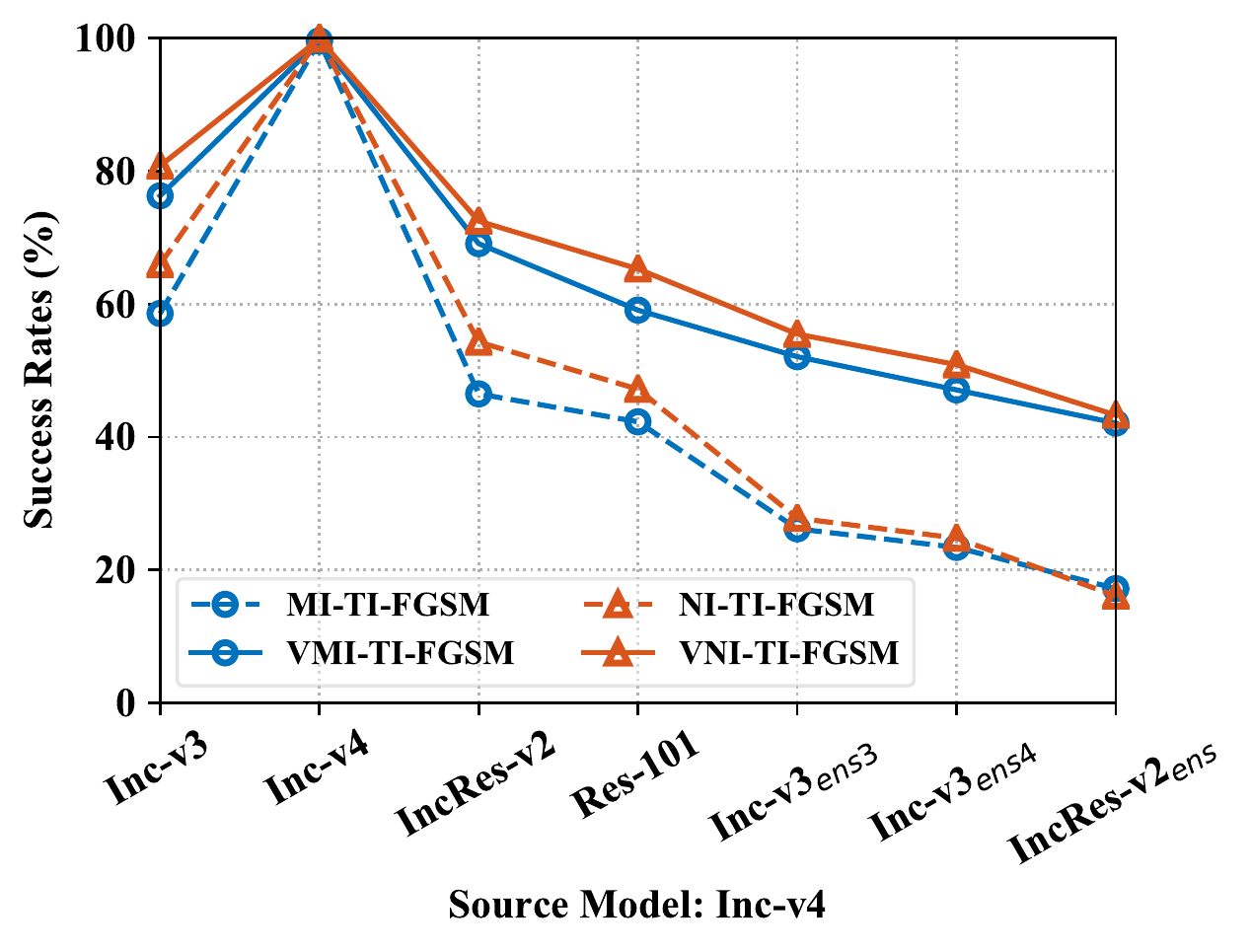}
        \end{subfigure}
        \begin{subfigure}{.33\textwidth} 
          \centering 
          \includegraphics[width=\linewidth]{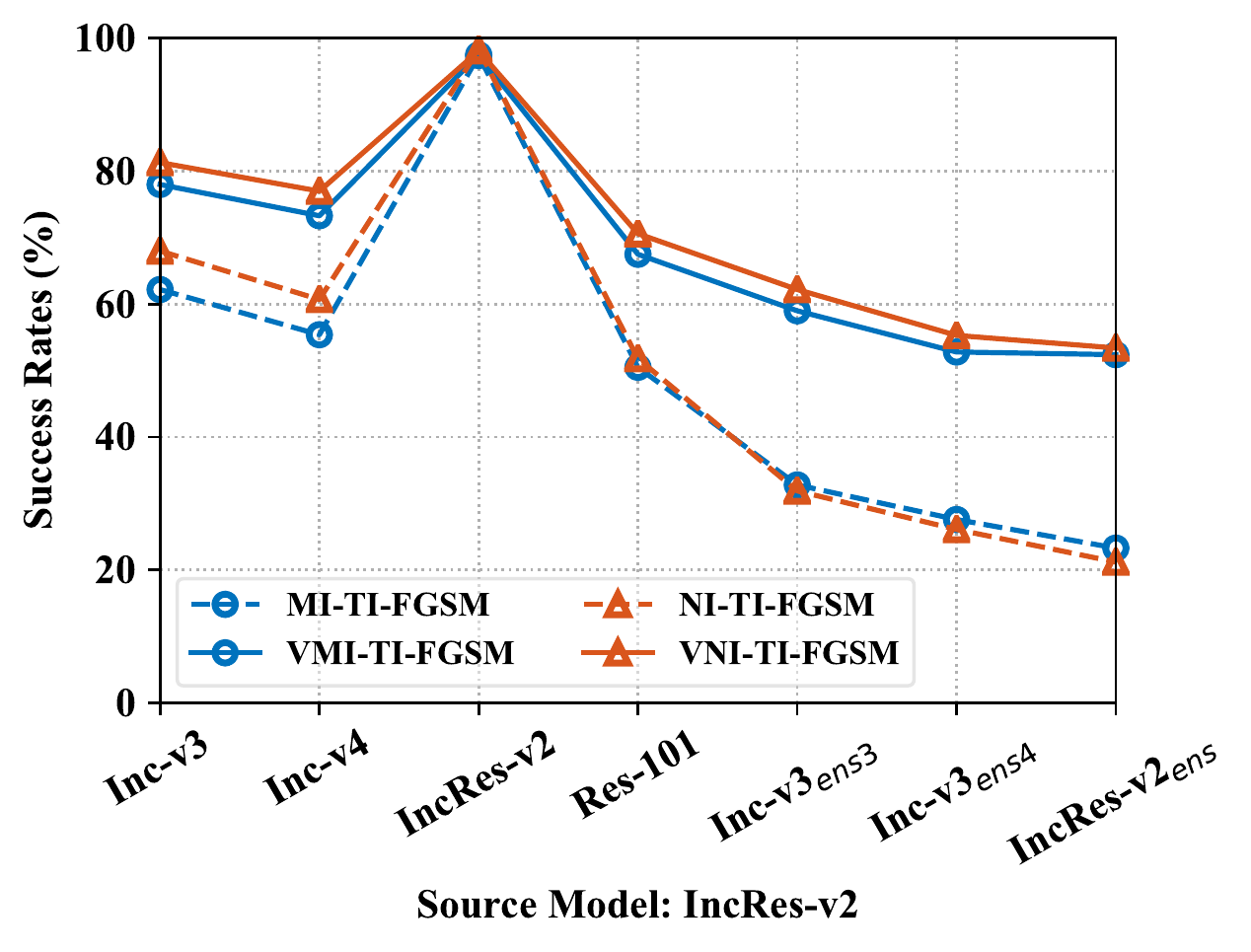}
        \end{subfigure}%
        \begin{subfigure}{.33\textwidth}
          \centering 
          \includegraphics[width=\linewidth]{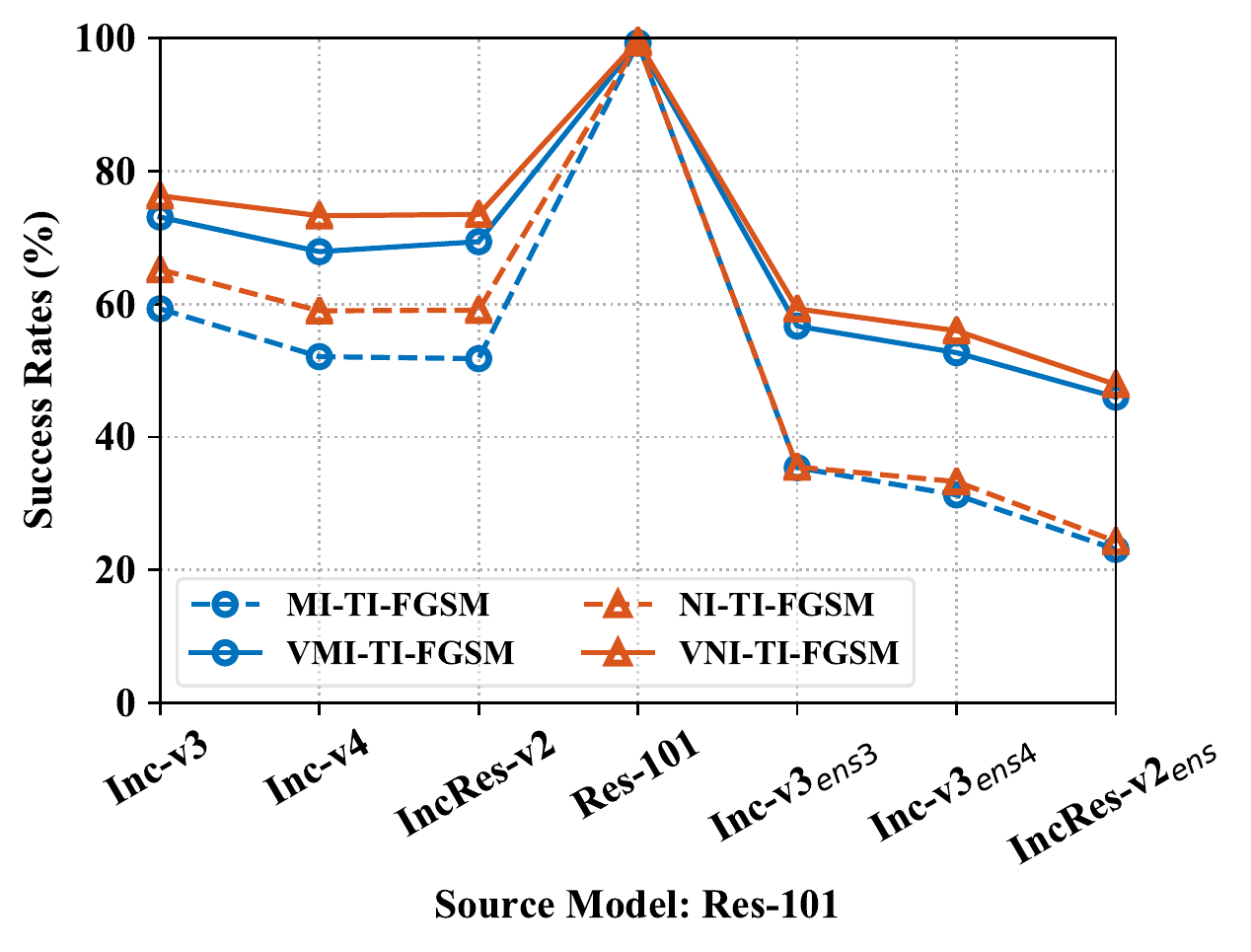}
        \end{subfigure}
    \caption{The success rates (\%) on seven models in the single model setting by various gradient-based iterative attacks enhanced by TIM.}
    \label{app:fig:tim}
    \vspace{1em}
    \centering
    \begin{subfigure}{.33\textwidth}
          \centering 
          \includegraphics[width=\linewidth]{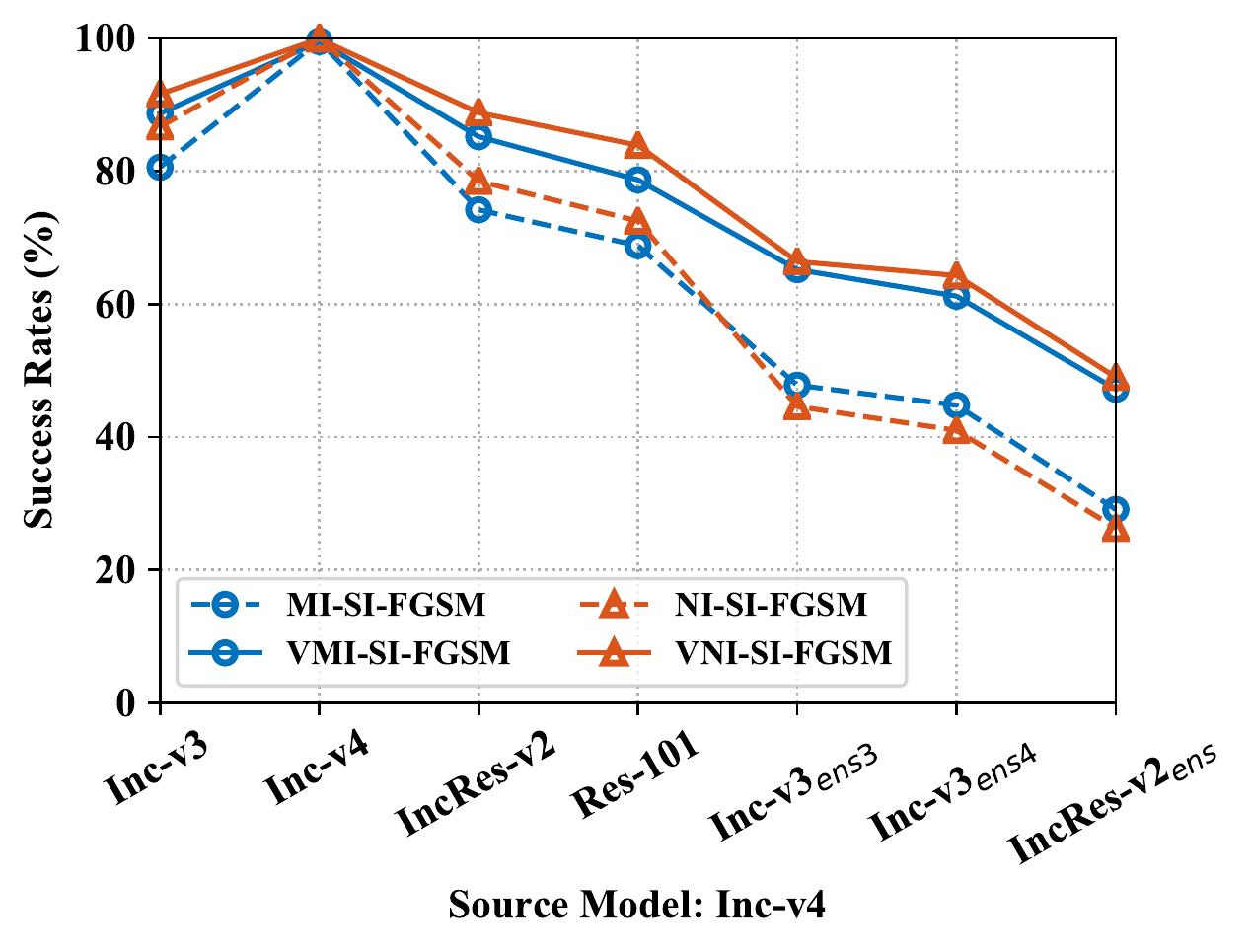}
        \end{subfigure}
        \begin{subfigure}{.33\textwidth} 
          \centering 
          \includegraphics[width=\linewidth]{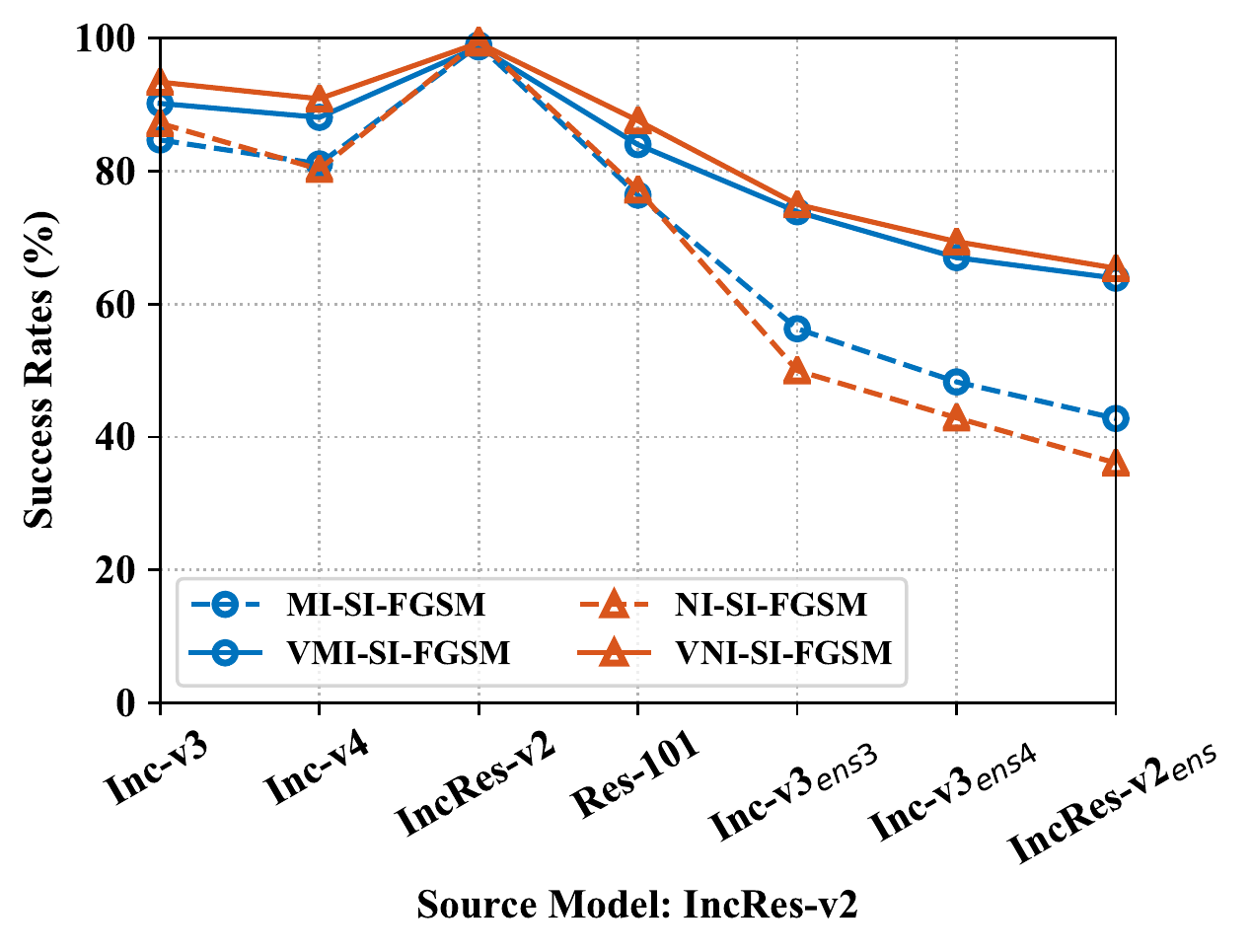}
        \end{subfigure}%
        \begin{subfigure}{.33\textwidth}
          \centering 
          \includegraphics[width=\linewidth]{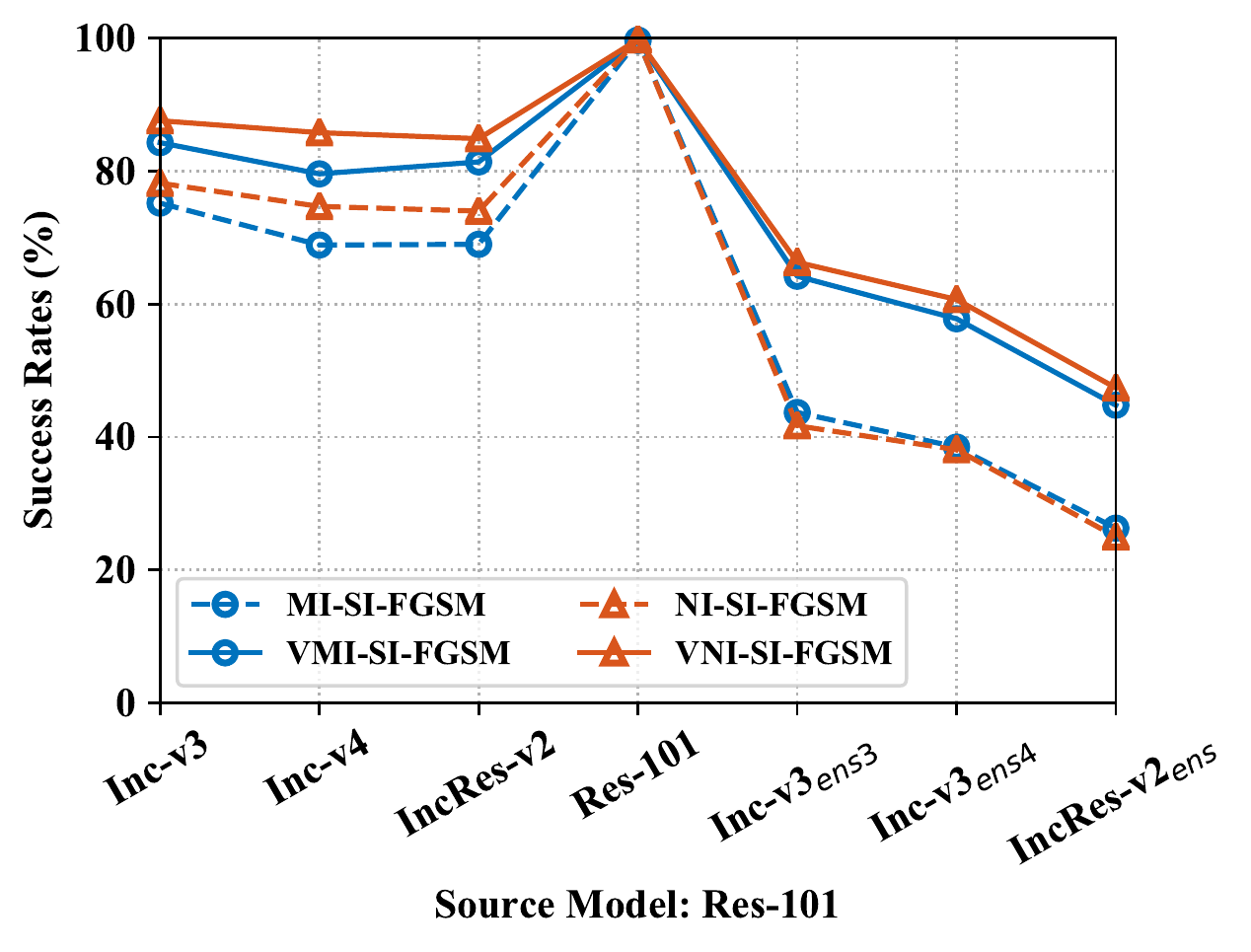}
        \end{subfigure}
    \caption{The success rates (\%)
    on seven models in the single model setting by various gradient-based iterative attacks enhanced by SIM.}
    \label{app:fig:sim}
\end{figure*}

Here we further provide the attack results of our methods with input transformations on the other three models, \ie Inc-v4, IncRes-v2 and Res-101. The results are depicted in Figure \ref{app:fig:dim} for DIM, Figure \ref{app:fig:tim} for TIM and Figure \ref{app:fig:sim} for SIM. Our methods can improve the transferability of these input transformations remarkably, especially against the adversarially trained models. The results are consistent with the results of adversarial examples crafted on Inc-v3 model.

\end{document}